\documentclass[10pt,twocolumn,letterpaper]{article}

\usepackage{iccv}
\usepackage{times}
\usepackage{epsfig}
\usepackage{graphicx}
\usepackage{subcaption}
\usepackage{amsmath}
\usepackage{amssymb}
\usepackage{multirow}
\usepackage{diagbox}
\usepackage{colortbl}
\usepackage{booktabs}
\usepackage{pifont}
\usepackage{makecell}
\usepackage{appendix}

\usepackage[pagebackref=true,breaklinks=true,letterpaper=true,colorlinks,bookmarks=false]{hyperref}

\iccvfinalcopy % *** Uncomment this line for the final submission

\ificcvfinal\fi

\begin{document}

%%%%%%%%% TITLE
\title{%Improved Few-shot Object Detection Utilizing Potential Objects [TO DO]
 Identification of Novel Classes for Improving Few-Shot Object Detection
}

\author{Zeyu Shangguan\\
University of Southern California\\
{\tt\small zshanggu@alumni.usc.edu}
\and
Mohammad Rostami\\
University of Southern California\\
{\tt\small rostamim@usc.edu}
}

\maketitle
\ificcvfinal\thispagestyle{empty}\fi

%%%%%%%%% ABSTRACT
\begin{abstract}
% Few-shot object detection (FSOD), as a way to realize robust object detection with only a few training samples available, has broad prospects for strong artificial intelligence (AI) applications and leading the future of AI to some extent.

Conventional training of deep neural networks requires a large number of the annotated image which is a laborious and time-consuming task, particularly for rare objects. Few-shot object detection (FSOD) methods offer a remedy by realizing robust object detection using only a few training samples per class. An unexplored challenge for FSOD is that instances from unlabeled novel classes that do not belong to the fixed set of training classes appear in the background. These objects behave similarly to label noise, leading to FSOD performance degradation. We develop a semi-supervised algorithm to detect and then utilize these unlabeled novel objects as positive samples during training to improve FSOD performance. Specifically, we propose a hierarchical ternary classification region proposal network (HTRPN) to localize the potential unlabeled novel objects and assign them new objectness labels. Our improved hierarchical sampling strategy for the region proposal network (RPN) also boosts the perception ability of the object detection model for large objects. Our experimental results indicate that our method is effective and outperforms the existing state-of-the-art (SOTA) FSOD methods.~\url{https://github.com/zshanggu/HTRPN} % on the PASCAL VOC and COCO datasets.
%However, a thorny problem is still hindering the development of FSOD: unlabelled novel objects that appear in the base classes data bring negative effects on the FSOD task.
\end{abstract}

%%%%%%%%% BODY TEXT
\section{Introduction}

The adoption of deep neural network architectures in object detection has led to a significant method in determining the location and the category of objects of interest in an image. In the presence of abundant training data, object detection models based on the region-based convolution neural networks (R-CNN) architecture reach high accuracy on most object detection tasks. However, preparing large-scale annotated training data can be a challenging task in some applications, e.g., miscellaneous disease analysis and industrial defect detection. In the presence of insufficient training data, these models easily overfit and fail to generalize well. In contrast, humans are able a novel object class very fast based on a few samples. As a result, it is extremely desirable to develop models that can learn object classes using only a few samples, known as few-shot object detection (FSOD).

\begin{figure}[t]
    \centering
    \includegraphics[width=80mm]{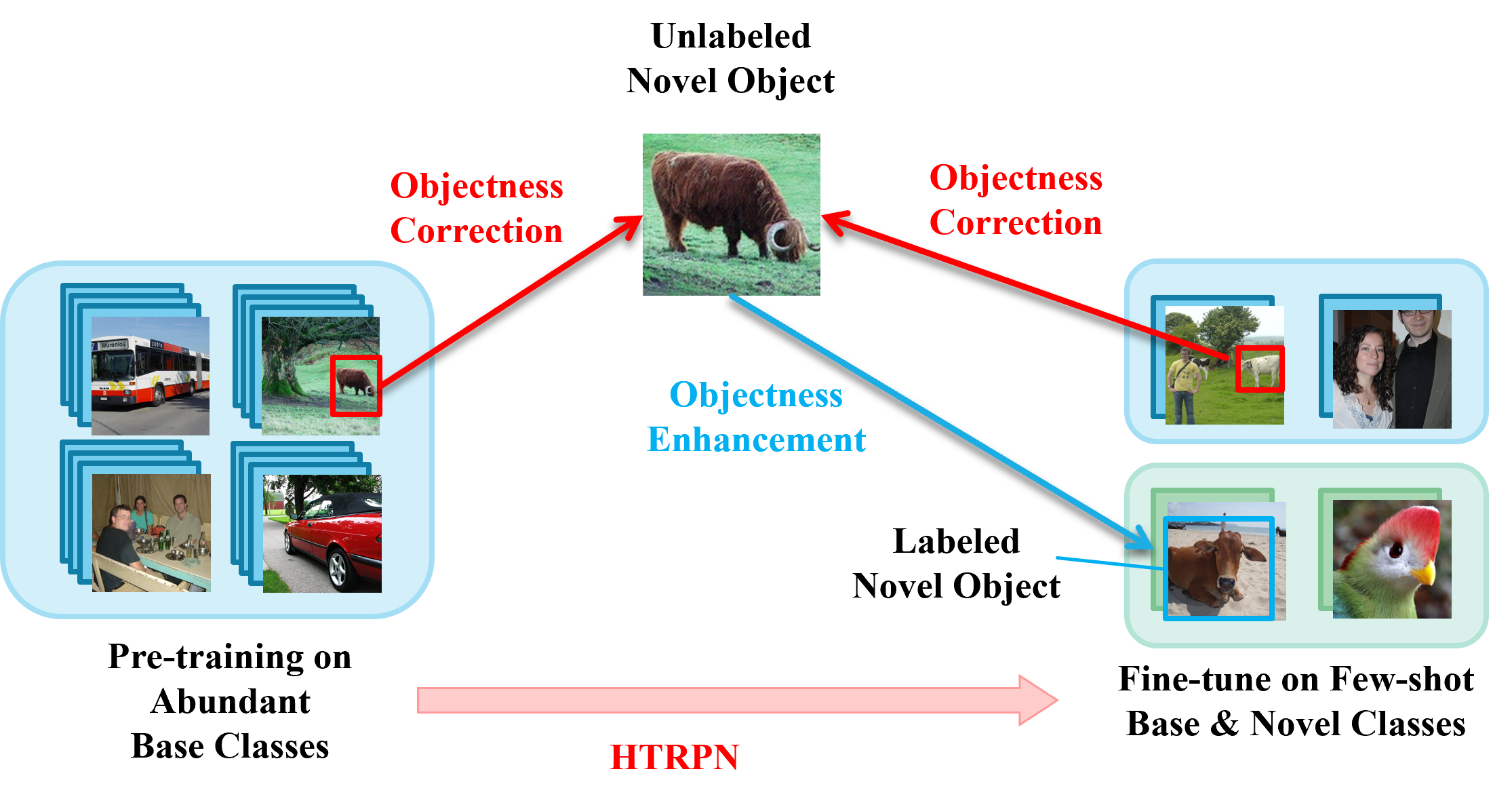}
    \caption{\small FSOD methods pre-train a model on abundant base classes and then fine-tune it on both base and novel classes.}
    \label{fig:intro}
\end{figure}

Current FSOD methods are based on pre-training a suitable model on a set of base classes with abundant training data and then fine-tuning the model on both the base classes and the novel classes for which only a few samples are accessible (see Figure~\ref{fig:intro}).     %Commonly used datasets that are used for evaluation, such as PASCAL VOC~\cite{Everingham10} and COCO~\cite{Lin14}, are divided into abundant base classes and few-shot novel classes. 
The primary approach in FSOD is to benefit from ideas in transfer learning or meta-learning to learn novel classes through the knowledge obtained during the pre-training stage while maintaining good performance in base classes. Despite recent advances in FSOD, current SOTA methods are still far from getting favorable results on novel classes similar to the base classes. Potential reasons for this performance gap include the confusion between visually similar categories, incorrect annotations (label noise), the existence of unseen novel objects during training, etc. Recent FSOD  methods have focused on addressing these challenges for improved FSOD performance.% on novel classes. %For example, we can see in Figure~\ref{fig:intro}  that some novel classes (e.g., the cow in the input image) might encounter objectness inconsistency, i.e., some of the instances remain unlabeled in the base class images and are therefore treated as background objects, while the labeled instances are treated as foreground objects. 

% Our proposed HTRPN applies a semi-supervised strategy to correct the objectness of the potential novel objects, consequently, the objectness inconsistency is eliminated

%Previous works have demonstrated meaningful opinions and novel benchmarks on these problems, but there is still a gap to catch up with the performance of the data-sufficient cases.

We study the phenomenon that unlabeled novel object classes that do not belong to either of the base or the labeled novel classes can appear in the training data. For example, we see in Figure.~\ref{fig:intro} that among base-class training samples, there are a number of objects that remain unlabeled, such as the cow in the image. These unlabeled objects can potentially belong to unseen novel classes. Our experiments demonstrate that this phenomenon exists in PASCAL VOC~\cite{Everingham10} and COCO~\cite{Lin14} datasets.% but also happens in real-life industrial applications. 
 % The following sentences are unclear to me
This phenomenon leads to the objectness inconsistency for the model when recognizing the novel objects: for the novel class, objects are treated as background if their annotations are missing, but they are treated as foreground where they are labeled. Such nonconformity of foreground and background confuses the model when training the objectness and make the model hard to converge and degrades detection accuracy.

To tackle the above challenge, we develop a semi-supervised learning method to utilize the potential novel objects that appear during training to improve the ability of the model to recognize novel classes. We first demonstrate the possibility of detecting these unlabeled objects. Our experiment indicates that some unlabeled class objects are likely to be recognized if they are similar to the training base and novel classes. We collect the unlabeled novel objects from the background proposals by determining whether they are predicted as known classes, and then we give these proposals an extra objectness label in the region proposal network (RPN) so that the model could learn them. We also analyze the defect of the standard RPN in detecting objects of different sizes during training and propose a more balanced RPN sampling method so that objects are treated equally in all scales. We provide extensive experimental results to demonstrate the effectiveness of our method on the PASCAL VOC and COCO datasets.
Our   contributions include:

\begin{itemize}
    \item We modify the anchor sampling strategy so that the anchors are evenly chosen from different layers of the feature pyramid layer in the R-CNN architecture. %Objects with relatively larger sizes have higher possibilities to be observed.
    
    \item We design a ternary objectness classification in the RPN layer which enables the model to recognize potential novel class objects to improve consistency. %The objectness consistent with the potential novel objects is well-improved.
    
    \item We use contrastive learning in the RPN layer to distinguish between the positive and the negative anchors.
\end{itemize}

\section{Related works}
%We survey works that are relevant to our idea for FSOD.

We assume that there are three classes: base classes, seen novel classes, and unseen novel classes. The base and the seen novel classes form the training dataset. For base classes, we have sufficient data but for seen novel classes, we have a few samples per class. Most works in FSOD only consider these classes. The unseen novel classes are not included in the training data but emerge as novel classes in the background. %We briefly survey relevant work.

\textbf{Few-shot object detection}
Typical object detection networks are usually either two-stage or one-stage. For two-stage object detection networks, such as R-CNN ~\cite{Girshick14}, R-FCN ~\cite{Dai16}, Fast-RCNN ~\cite{Girshick15}, and Faster-RCNN ~\cite{Ren15}, the model first applies fixed anchors in the region proposal network (RPN) to determine if a proposal box contains an object. The selected proposals then are sent to the region of interest (RoI) pooling layer to get an instance-level classification and bounding boxes. One-stage object detection networks such as SSD ~\cite{Liu16}, YOLO series ~\cite{Redmon16, Redmon18}, and Overfeat ~\cite{Sermanet14}, estimate the category and the location of an object directly from the backbone network without RPN. Two-stage object detection networks have higher detecting accuracy than one-stage schemes but lower inference speed ~\cite{Kohler22}.
Few-shot object detection (FSOD), is a case of object detection that only a few samples are available for training.%, is much closer to human learning. Humans could learn from very few examples since we have rich prior experiences. Recent research on few-shot learning has demonstrated that sufficient pre-training is non-negligible and could significantly improve the recognition ability of the model when transferring to a new few-shot task, especially with huge models such as vision transformer~\cite{Hu22}.

\textbf{Two-stage FSOD}
For FSOD,  the model is usually first pre-trained on the base classes for which we have data-sufficient. The model is then, fine-tuned on the seen novel classes ~\cite{Wang20}, each with a few samples. As for the fine-tuning stage, meta-learning and transfer learning are two major end-to-end approaches. Methods based on meta-learning~\cite{Fan20, Han22a, Han22b, Yin22} build an inquiry set and a support set that contains $k$ categories with $n$ samples in each, namely $k$-way $n$-shot setting. By creating the $k$-way $n$-shot episodes for training, meta-learning help to learn a metric to determine which support set category an inquiry image belongs to~\cite{Fan20, Han22a, Han22b, Yin22}. % such works include Meta Faster R-CNN. 
In contrast, methods based on transfer learning start from the pre-training weights and fine-tune the model on the novel seen classes ~\cite{Wang20, Sun21}.

\textbf{Unseen novel objects}
In an object detection problem, the set of the base and the seen novel classes are assumed to be a closed set. However, there may be potential novel unseen objects in the training data that do not belong to the initial set of classes. These objects naturally are classified as one of seen classes and hence, there has been an interest to mitigate the adverse effects of these objects. Semi-supervised object detection network is a potential solution for this problem which utilizes the challenging samples ~\cite{Rosenberg05, Liu21, Xu21}. Kaul et al. ~\cite{Kaul22} build a class-specific self-supervised label verification model to identify candidates of unlabeled (unseen) objects and give them pseudo-annotations. The model is then retrained with these pseudo-annotated samples to improve the object-detecting accuracy. However, this method requires two rounds of training and requires extra effort to adapt to other categories. Li et al. ~\cite{Li21c} propose a distractor utilization loss by giving the distractor proposals a pseudo-label during fine-tuning. This method is used only in the fine-tuning stage and hence, the objectness inconsistency from the pre-training stage is not addressed. Inspired by these shortcomings, we propose utilizing the unlabeled potential objects that belong to the unseen classes to reduce the negative effect of novel objects.

\textbf{Contrastive learning}
 can be used to enlarge the inter-class distances and narrow down the intra-class distances for classification tasks to enhance data representations. It has been applied to many classification tasks in topics such as visual recognition ~\cite{Luo21, Wang21a}, semantic segmentation ~\cite{Wang21b}, super-resolution ~\cite{Wang23}, and natural language processing ~\cite{Fang05, Chi20}. Self-supervised contrastive learning in few-shot object detection is introduced by FSCE~\cite{Sun21} to better distinguish similar categories at the instance level during fine-tuning. We benefit from a similar strategy in our work.% to improve the objectness evaluation ability of the RPN layer. %FSCE provides a   baseline network for  FSOD. We improve upon FSCE by improving the objectness evaluation ability of the RPN layer.

\begin{figure}[t]
	\centering
	\begin{subfigure}{0.45\linewidth}
		\centering
		\includegraphics[width=0.8\linewidth]{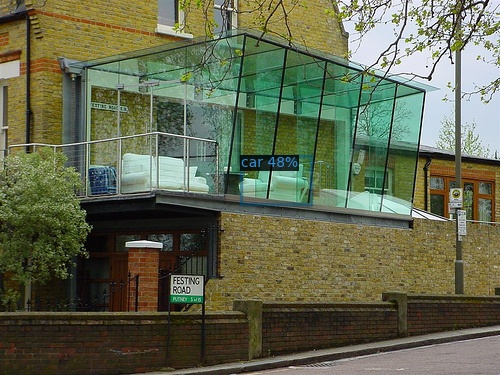}
		\caption{}
		\label{figure2_a}
	\end{subfigure}
	\centering
	\begin{subfigure}{0.45\linewidth}
		\centering
		\includegraphics[width=0.8\linewidth]{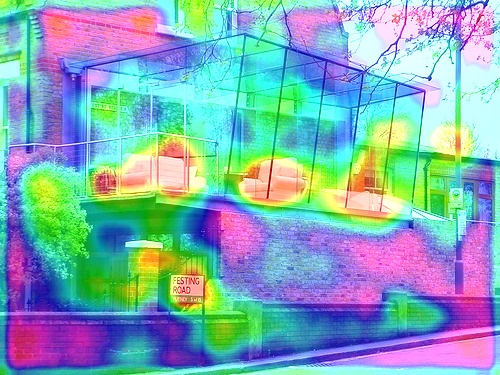}
		\caption{}
		\label{figure2_b}
	\end{subfigure}
	\centering
	\begin{subfigure}{0.45\linewidth}
		\centering
		\includegraphics[width=0.8\linewidth]{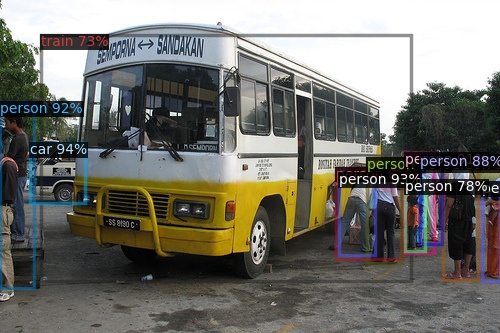}
		\caption{}
		\label{figure2_c}
	\end{subfigure}
	\centering
	\begin{subfigure}{0.45\linewidth}
		\centering
		\includegraphics[width=0.8\linewidth]{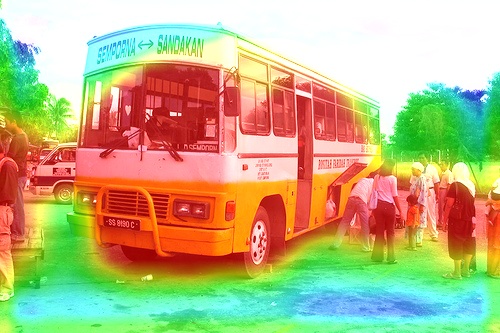}
		\caption{}
		\label{figure2_d}
	\end{subfigure}
	\caption{The left images in each row indicate the predicted boxes with the class names and confidence, and the right images are the feature maps of the RPN layer.}
	\label{fig:visualization}
\end{figure}

\section{Problem Description}
\label{sec: Problem Description}

We formulate the problem of FSOD following a standard setting in the literature~\cite{Kang19, Wang20}. We use the Faster R-CNN network as the object detection model and follow the same evaluation paradigm defined by ~\cite{Wang20}. Accordingly, the base classes are those classes for which we have sufficient images and instances for each base class ($C_B$), while novel classes are those for which, we only have a few training samples in the dataset ($C_N$), where $C_B\cap C_N = \varnothing$. An $n$-shot learning scenario means that we have access to $n$ images per seen novel categories. During the pre-training stage, the model is trained only on base class $C_B$, and also is only evaluated on the test set of $C_B$. We then proceed to learn the seen novel classes in the second stage. To overcome catastrophic forgetting about the learned knowledge about the base classes, the pre-trained model is then fine-tuned on both the seen novel classes and base classes $C_N\cup C_B$ and then is tested on both sets of classes.

Architectures based on R-CNN have been used consistently for object detection. In our work, we improve the R-CNN architecture to identify novel unseen classes as instances that do not belong to the seen classes.
For an input image, R-CNN derives five scaled feature maps ($p2\sim p6$) using its feature pyramid network (FPN), and then size-fixed anchors in the region proposal network (RPN) are applied on these feature maps to predict the objectness (i.e., $obj\{obj_{pre}, obj_{gt}, iou_{gt}^a\}$, where $obj_{pre}$ is the predicted objectness score in the range of 0 and 1. Here, the ground truth value $obj_{gt}=0$ indicates a non-object and $obj_{gt}=1$ represents a true object, and $iou_{gt}^a$ represents the intersection over the union of an anchor with its ground truth box) and the coarse bounding box (i.e., $bbox_c$) of each anchor to get proposal boxes (i.e., $Prop\{obj, bbox_c, iou_{gt}^p\}$, $iou_{gt}^p$ represents the intersection over the union of a proposal box with its ground truth box). Anchors with $iou_{gt}^a > 0.7$ are called active anchors ($A_a$) and their corresponding proposals are called positive proposals ($Prop_p$); while anchors with $iou_{gt}^a < 0.3$ are called negative anchors ($A_n$) and the corresponding proposals are called negative proposals ($Prop_n$). Next, $Prop_p$ and $Prop_n$ are sent to the region of interest pooling layer (RoI pooling) to predict their instance level classification (i.e., $cls^i$, where $i$ is the classification index) and the refined bounding box (i.e., $bbox_r$). The objects ($Obj\{cls^i, bbox_r\}$) are then finally detected.

\begin{figure}[t]
	\centering
    \begin{subfigure}{\linewidth}
        \centering
    	\includegraphics[width=30mm]{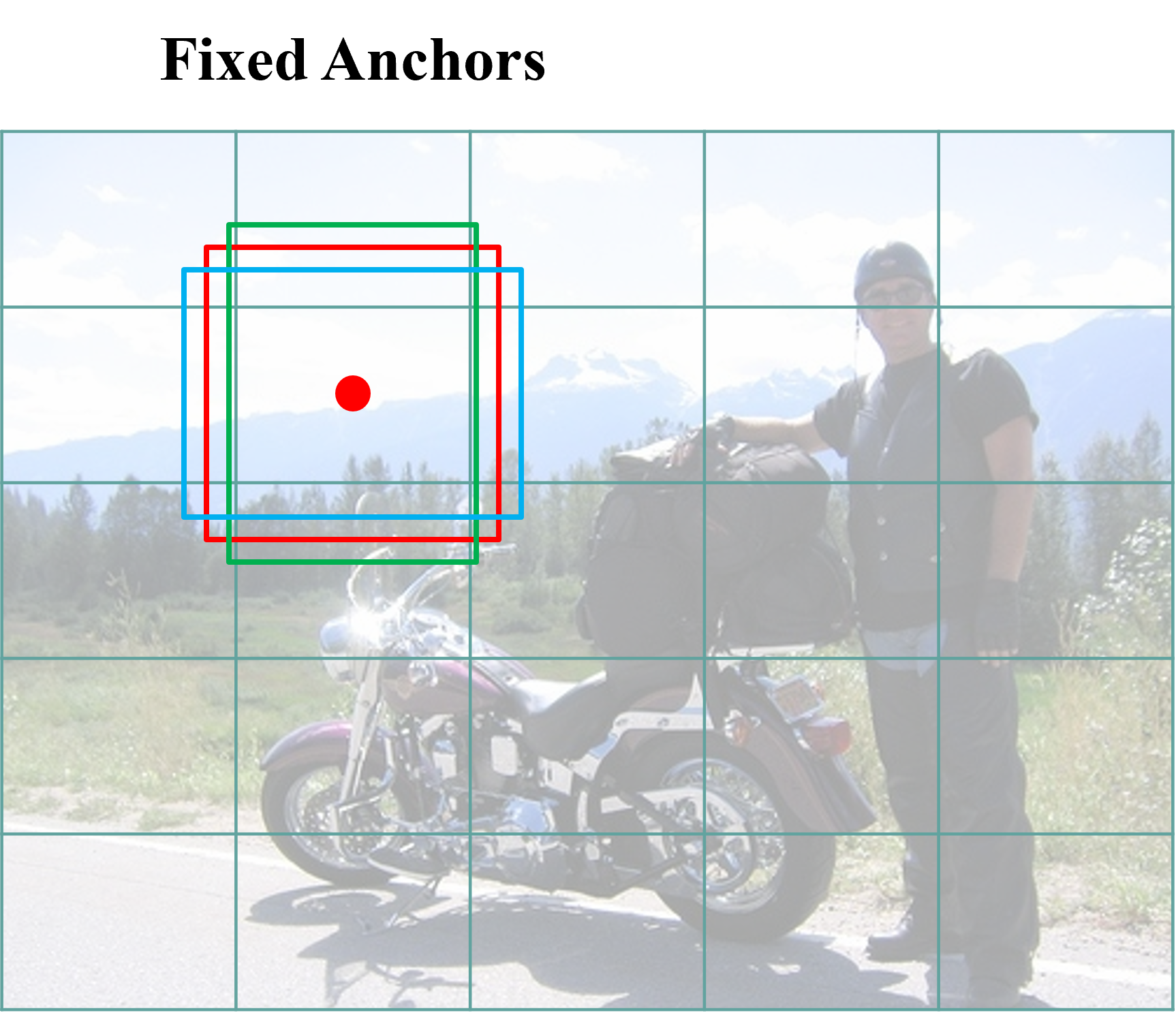}
    	\caption{}
    	\label{fig:anchors}
    \end{subfigure}
    \centering
    \begin{subfigure}{\linewidth}
        \centering
    	\includegraphics[width=85mm]{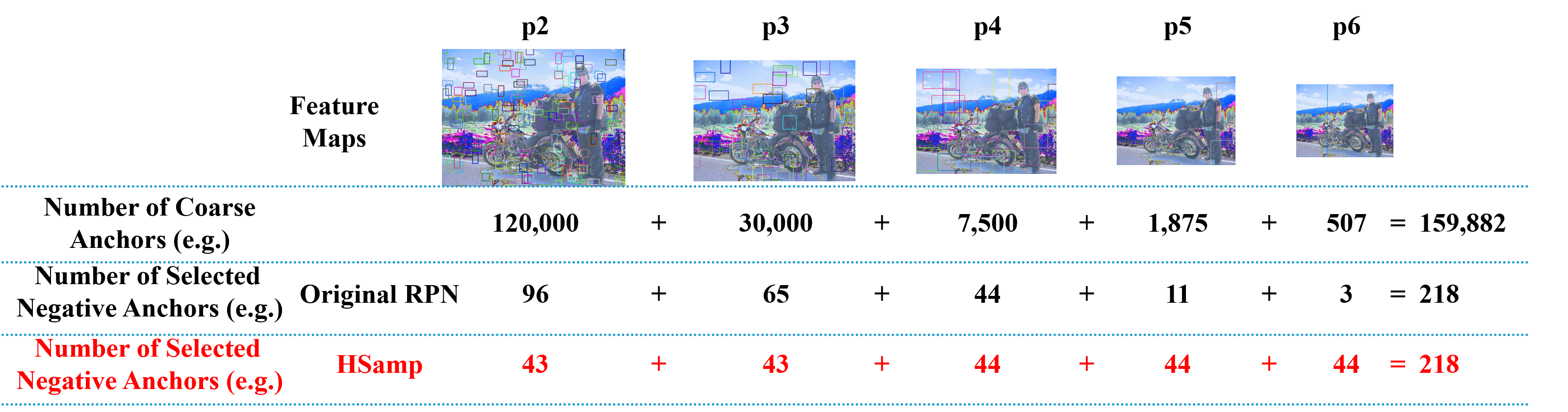}
    	\caption{}
    	\label{fig:HirecRPN}
    \end{subfigure}
	\caption{ Novel unseen classes versus novel seen classes: (a) a schematic diagram of how coarse anchors work. For an $n\times m$ feature map, 3 fixed anchors will be applied on each pixel of it. Therefore, each feature map would have $3m\times n$ coarse anchors. (b) The mechanism of our proposed HSamp can equally choose anchors for each layer.}
	\label{fig:HSamp}
\end{figure}

\begin{figure*}[htbp]
	\centering
	\begin{subfigure}{0.15\linewidth}
		\centering
		\includegraphics[width=0.9\linewidth]{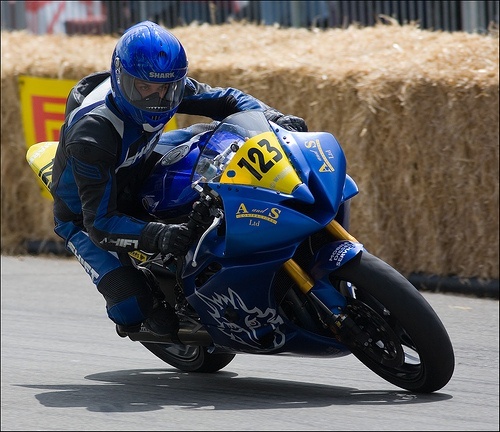}
		\caption{}
		\label{figure6_ori_a1}
	\end{subfigure}
	\centering
	\begin{subfigure}{0.15\linewidth}
		\centering
		\includegraphics[width=0.9\linewidth]{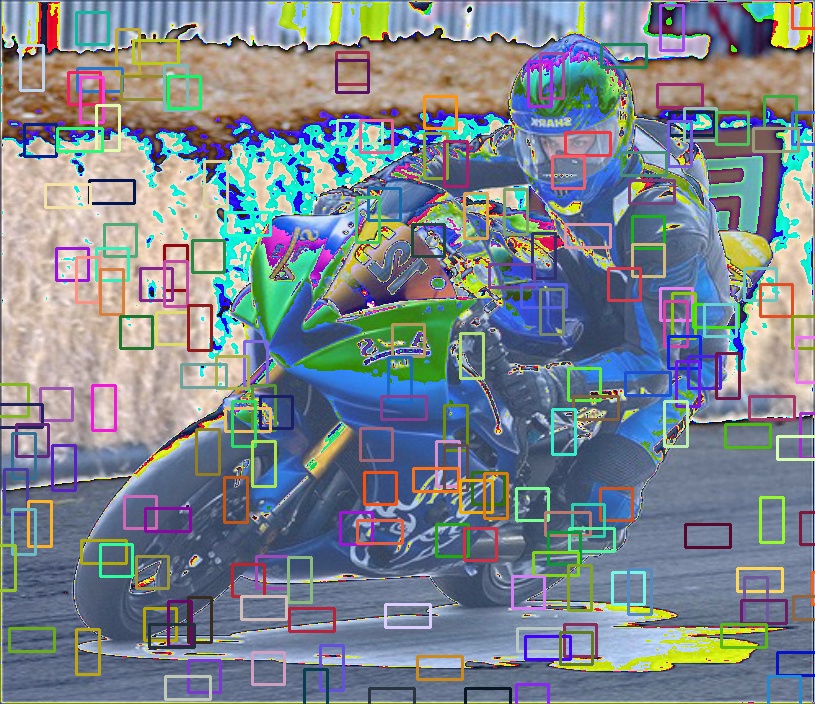}
		\caption{}
		\label{figure6_ori_a2}
	\end{subfigure}
	\centering
	\begin{subfigure}{0.15\linewidth}
		\centering
		\includegraphics[width=0.9\linewidth]{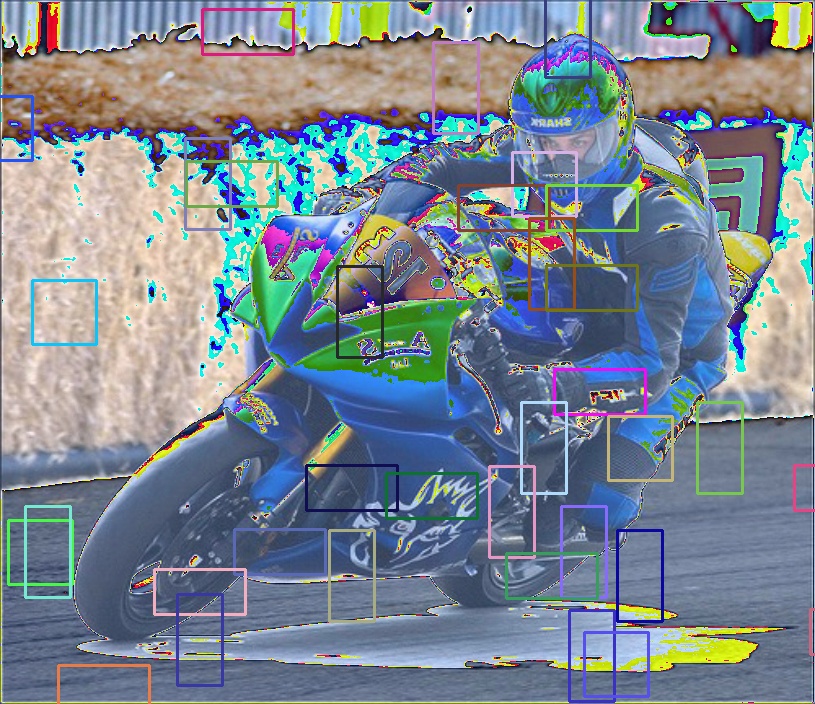}
		\caption{}
		\label{figure6_ori_a3}
	\end{subfigure}
	\centering
	\begin{subfigure}{0.15\linewidth}
		\centering
		\includegraphics[width=0.9\linewidth]{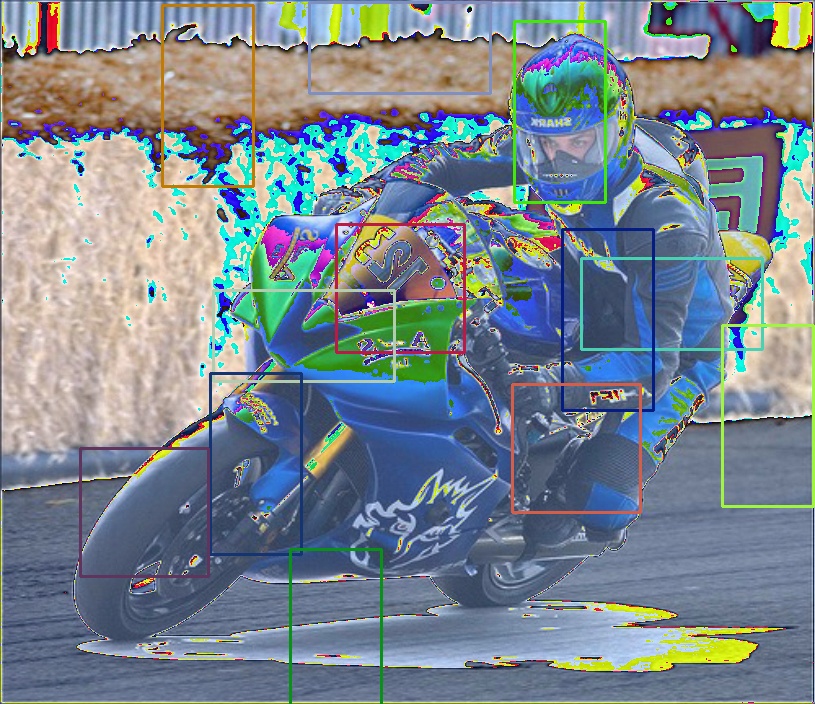}
		\caption{}
		\label{figure6_ori_a4}
	\end{subfigure}
	\centering
	\begin{subfigure}{0.15\linewidth}
		\centering
		\includegraphics[width=0.9\linewidth]{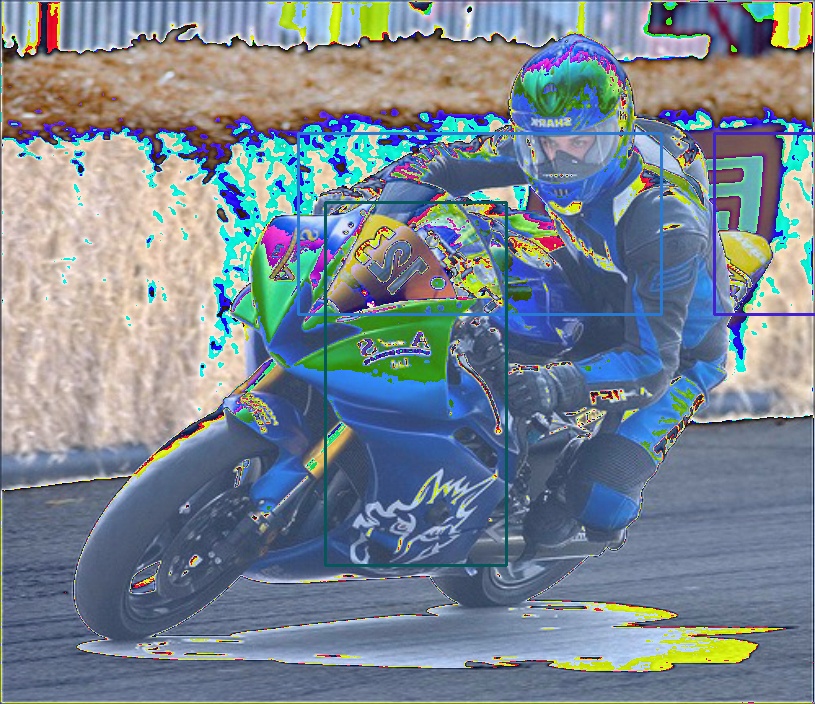}
		\caption{}
		\label{figure6_ori_a5}
	\end{subfigure}
	\centering
	\begin{subfigure}{0.15\linewidth}
		\centering
		\includegraphics[width=0.9\linewidth]{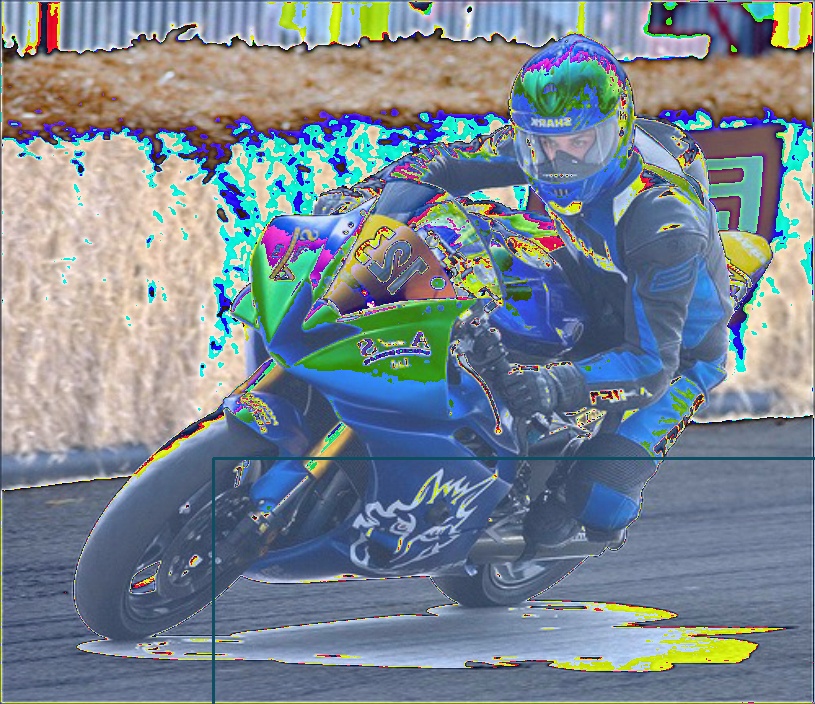}
		\caption{}
		\label{figure6_ori_a6}
	\end{subfigure}
        \centering
	\begin{subfigure}{0.15\linewidth}
		\centering
		\includegraphics[width=0.9\linewidth]{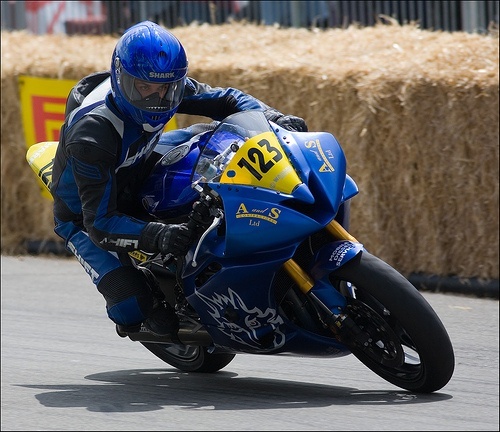}
		\caption{}
		\label{figure6_new_a1}
	\end{subfigure}
	\centering
	\begin{subfigure}{0.15\linewidth}
		\centering
		\includegraphics[width=0.9\linewidth]{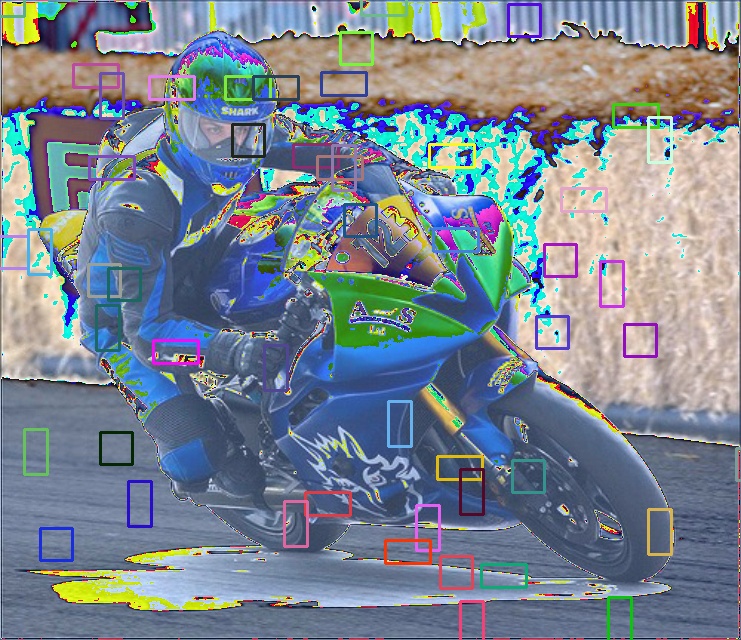}
		\caption{}
		\label{figure6_new_a2}
	\end{subfigure}
	\centering
	\begin{subfigure}{0.15\linewidth}
		\centering
		\includegraphics[width=0.9\linewidth]{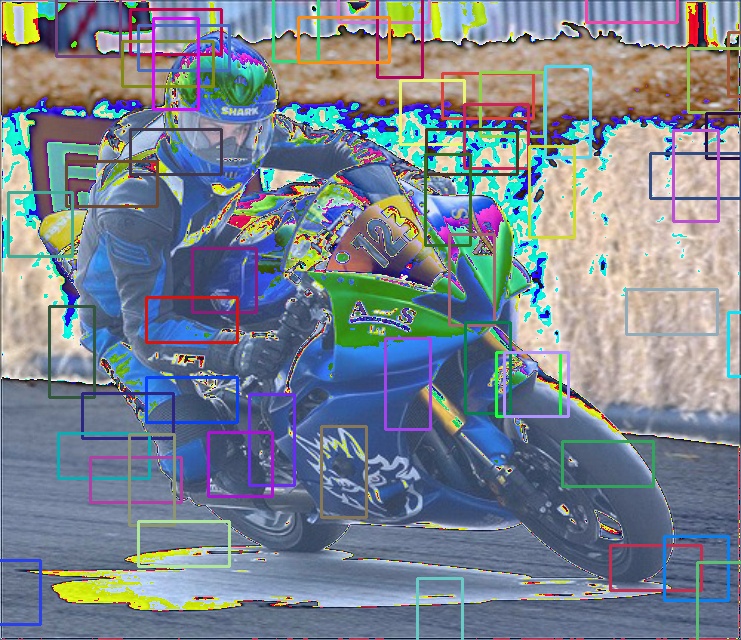}
		\caption{}
		\label{figure6_new_a3}
	\end{subfigure}
	\centering
	\begin{subfigure}{0.15\linewidth}
		\centering
		\includegraphics[width=0.9\linewidth]{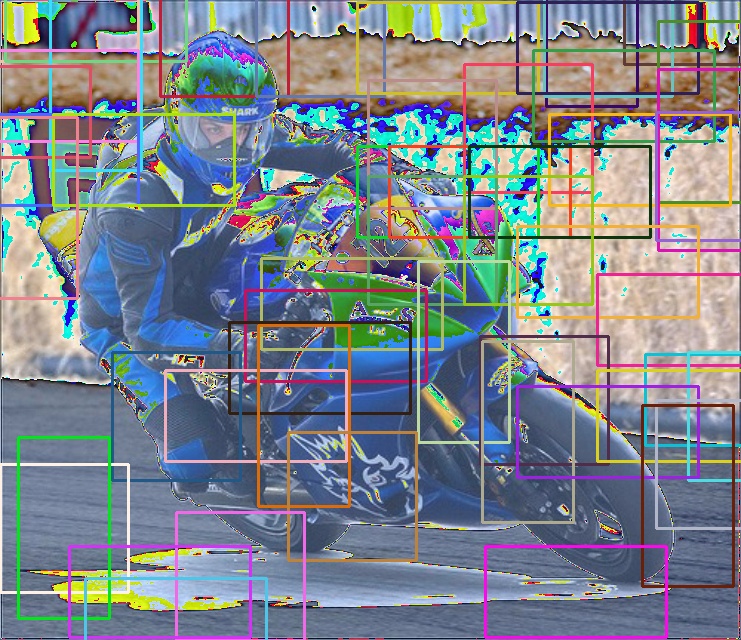}
		\caption{}
		\label{figure6_new_a4}
	\end{subfigure}
	\centering
	\begin{subfigure}{0.15\linewidth}
		\centering
		\includegraphics[width=0.9\linewidth]{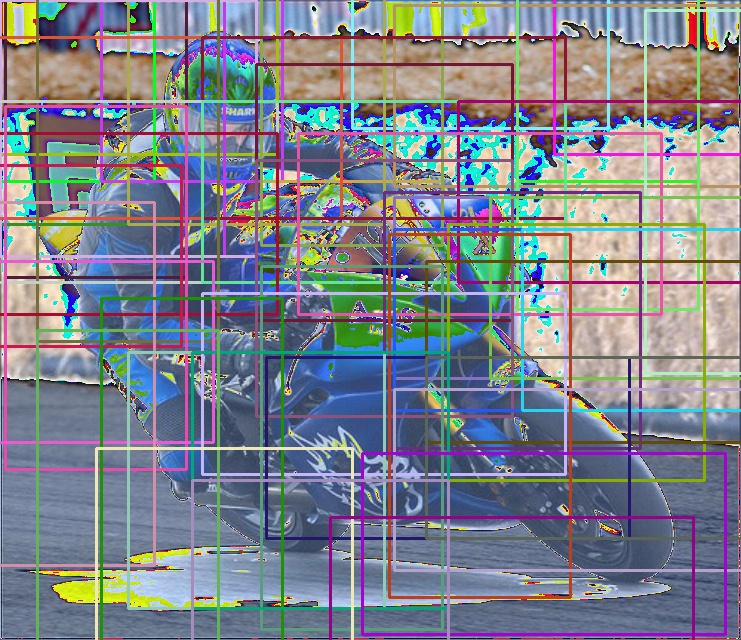}
		\caption{}
		\label{figure6_new_a5}
	\end{subfigure}
	\centering
	\begin{subfigure}{0.15\linewidth}
		\centering
		\includegraphics[width=0.9\linewidth]{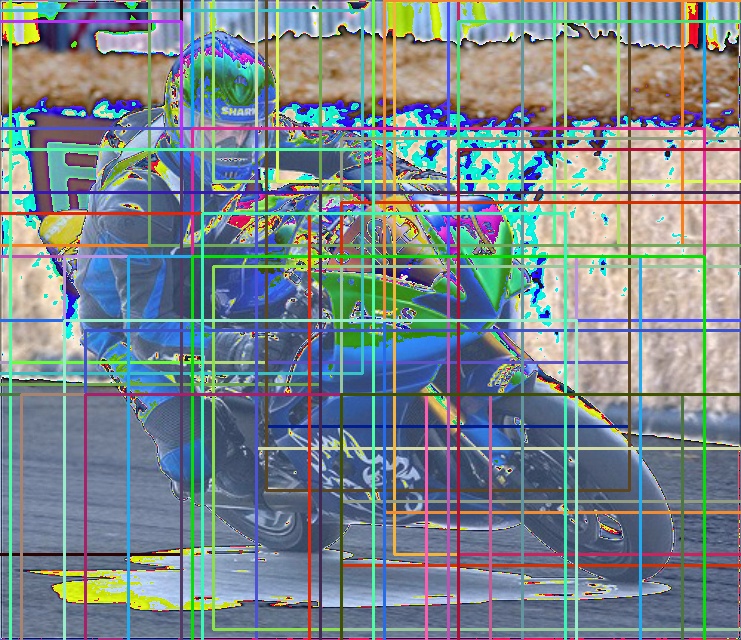}
		\caption{}
		\label{figure6_new_a6}
	\end{subfigure}
	\caption{The original image and anchors of p2/p3/p4/p5/p6 layers: (a)-(f) in the original FRCN, we hardly have chance to pick up negative anchors that contain large novel unseen class objects. (g)-(l) when HSamp is used, the anchors are more balanced for each layer and the large objects are more likely to be contained.}
	\label{fig:visualized HSamp}
\end{figure*}

The challenge that we want to address stems from the fact that
  an instance from the unlabeled and unseen novel object (${C_N}^{pn}$) can appear in the training dataset in the background (see Figure~\ref{fig:visualization}). The reason is that there are many potential classes that we have not included in either the base classes or the unseen novel classes. When detected, these objects would be treated as $Prop_n$ and with its ground truth objectness ${obj_{gt}}^{pn}=0$. On the contrary, they would be treated as $Prop_p$ with ground truth objectness ${obj_{gt}}^{pn}=1$ if it is labeled as such by the model. These instances can significantly confuse the model when adapting the model for learning the novel unseen classes. We argue that if the unlabeled potential novel object can be distinguished from the $Prop_n$, then its objectness could be modified as a foreground object. Consequently, the inconsistency of the objectness would be eliminated.
In other words, we propose to reduce an effect similar to noisy labels as these objects would be objects with wrong labels, leading to confusion in the model and performance degradation.

\section{Proposed solution}
\label{sec: Proposed Solution}
We outline our solution in this section. We first demonstrate that it is possible to encounter instances of novel unseen classes during the training stage. We then investigate the relationship between the number of anchors and the size of objects in each feature layer and provide a more effective sampling method for object detection. Finally, we describe our proposed pipeline to pick up objects from unseen classes with high confidence and then explain how we can modify their objectness loss to reduce their adverse effect.

\subsection{Finding the Potential Proposals}
\label{sec: Find the potential proposals}

According to the original Faster R-CNN network~\cite{Ren15}, the unlabeled area in an image would be wrongly treated as background objects in the RPN layer during training. Therefore, the potential unlabeled objects from unseen novel classes are suppressed and hard to be identified as true objects. However, to correct the objectness of these potential objects, the first step is how to find them. We have observed the fact that the network often has clear attention to the potential objects in the RPN layer. As an example, we have used Grad-CAM visualization of the feature map of the RPN layer on some representative training images, as shown in Figure.~\ref{fig:visualization}. We observe that although unlabelled novel objects appear in the base training images, the RPN layer could still have strong attention to them, and consequently predict some of them as a known class. In Figure.~\ref{figure2_a} and ~\ref{figure2_b}, the feature map of $p3$ layer clearly shows the attention of the ``chair'' (base class) and ``sofa'' (novel class), but the sofa is predicted to be an instance of the base class ``car''. Similarly, in Figure.~\ref{figure2_c} and ~\ref{figure2_d}, the potential novel objects (``bus'') can also be seen in the feature map of $p4$ layer, and the ``bus'' is predicted as an instance of the base class ``train''. This observation serves as an inspiration to identify potential proposals: some novel objects have high possibilities to be predicted as known base class.

Potential novel unseen class objects that appear in the base class training images usually have lower $iou_{gt}^a$ ~\cite{Li21c} and therefore must be contained by negative anchors ($A_n$). Theoretically, there are always exists anchors that can include the potential novel objects in an image. According to the architecture of the RPN layer, anchor boxes are used to determine whether an area contains objects, and each pixel of the feature map will have 3 fixed anchors that are in different sizes and aspect ratios, as shown in Figure.~\ref{fig:anchors}. Consequently, the overall number of anchors decreases for higher feature maps. In the original RPN layer, different sizes of anchors are applied according to the size of the $p2$ to $p6$ feature maps. Large anchors are more suitable for detecting large objects in high feature layers due to having a larger receptive field, and vice versa. Based on this inch-by-inch sliding window liked search, there should exist a sufficient number of candidates $A_n$ such that they contain potential unseen novel objects.
  % For a better illustration, we defined all the original negative anchors as coarse negative anchors ($A_{CN}$); within $A_{CN}$, anchors contains potential novel objects are called potential novel anchors ($A_{PN}$), and others are true negative anchors ($A_{TN}$). The relationship is $A_{TN}=A_{PN}\cup A_{TN}$. Accordingly, the positive anchors are $A_P$.
 However, to improve the training speed, not all of the anchors are used for determining proposal boxes. For an image, only 256 $A_a$ and $A_n$ anchors among all feature maps are randomly chosen to participate in the RoI pooling. Nonetheless, this random selection process dramatically reduces the chance to get desired negative anchors for large objects in higher feature maps in terms of probability, since the anchors of large size in $p4$ to $p6$ layers intrinsically have fewer cardinal numbers.

\begin{figure*}[ht]
	\hfill
	\begin{subfigure}{0.9\linewidth}
		\centering
		% \fbox{\rule{0pt}{2in} \rule{.9\linewidth}{0pt}}
		\includegraphics[width=150mm]{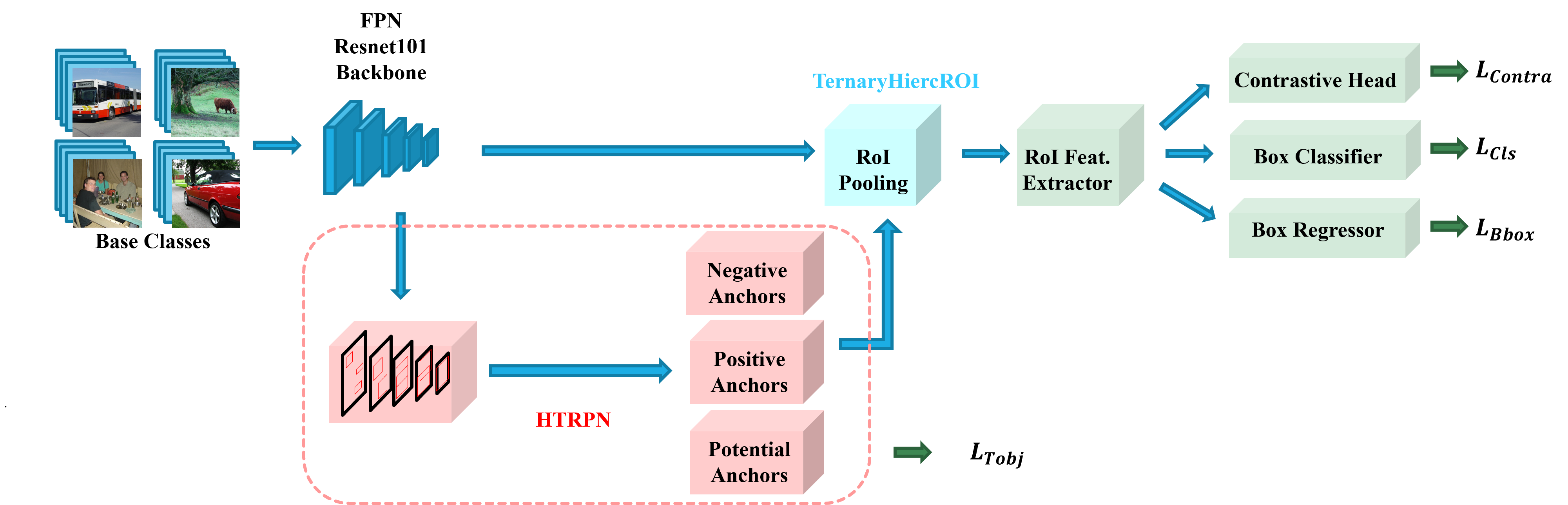}
		\caption{End-to-end architecture.}
		\label{fig:flowchart-a}
	\end{subfigure}
	\hfill
	\begin{subfigure}{0.9\linewidth}
		\centering
		% \fbox{\rule{0pt}{2in} \rule{.9\linewidth}{0pt}}
		\includegraphics[width=150mm]{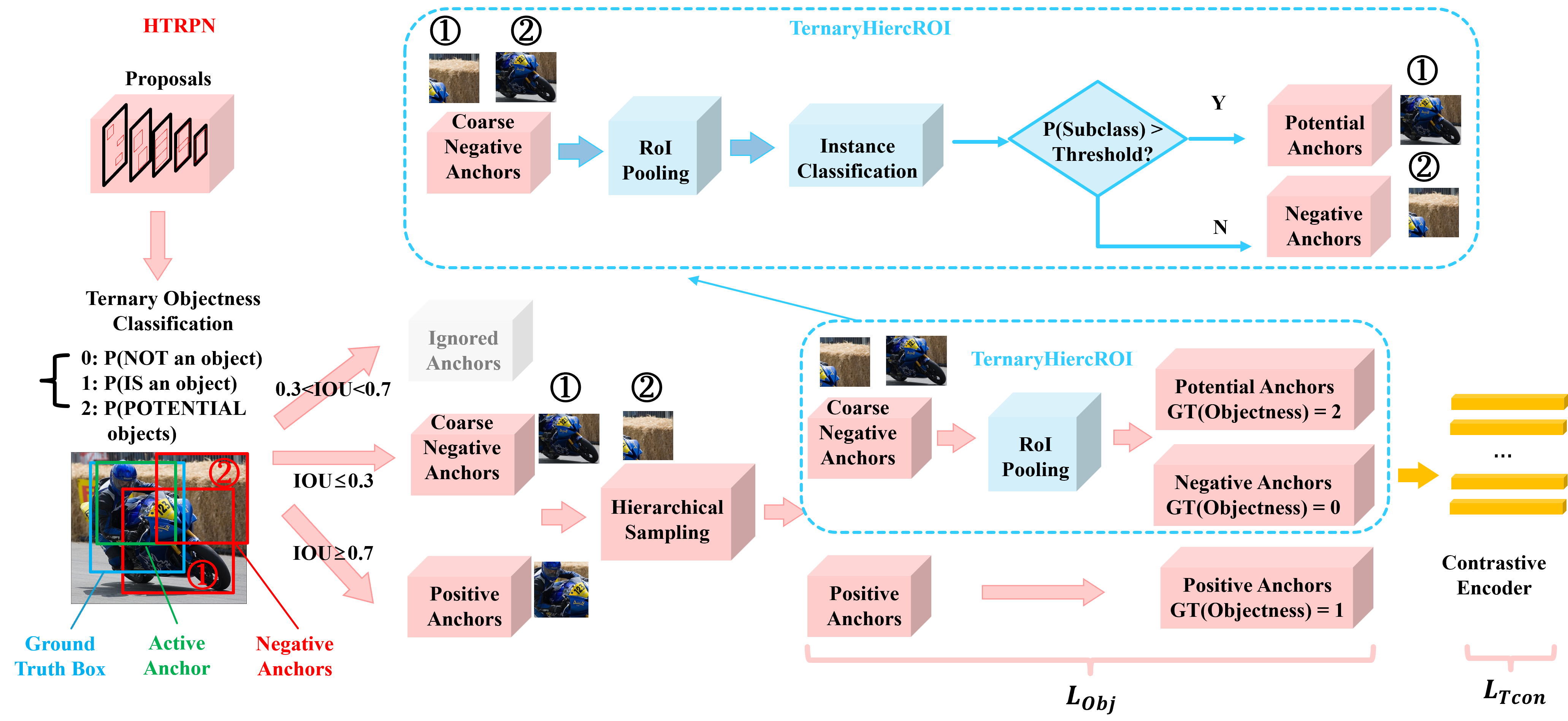}
		\caption{Detailed description of our proposed TernaryHiercRPN.}
		\label{fig:flowchart-b}
	\end{subfigure}
	\caption{Our proposed TernaryHiercRPN architecture.}
	\label{fig:flowchart}
\end{figure*}

To identify instances of novel unseen classes, we randomly select $A_n$s in a hierarchically balanced way, namely hierarchically sampling (HSamp). That is, if we need to pick up $m$ negative anchors ($m\textless 256$), we equally assign them to each feature layer so that each layer will have around $m/5$ anchors. This way, the anchors in each feature layer would share the same possibility for being selected, as shown in Figure.~\ref{fig:HirecRPN}. Therefore, the anchors that belong to the $p4$ to $p6$ layers are safely preserved. For example, in Figure.~\ref{fig:HirecRPN}, there are [[120,000], [30,000], [7,500], [1,875], [507]] anchors for $p4$ to $p6$ layers in a training batch, and 218 negative anchors are needed. With the original RPN, these 218 $A_n$ anchors are randomly selected which means the number of $A_n$ for $p6$ feature map is only 3. In contrast, when HSamp is used, the number of $A_n$ is equal for each feature map. We have also visualized the effect of our method in Figure~\ref{fig:visualized HSamp}. There is hardly any $A_n$ that contains the motorbike (novel unseen object) when using the original RPN, as shown in Figure.~\ref{figure6_ori_a1} to~\ref{figure6_ori_a6}. However, when HSamp is used, the chance to have an $A_n$ that contains the motorbike is higher, as shown in Figure.~\ref{figure6_new_a1} to~\ref{figure6_new_a6}. We conclude that it is crucial to implement a balanced strategy in sampling negative anchors among all feature layers in order to find potential objects that belong to unseen novel classes. The approach will help us to isolate unseen novel class instances as instances that are not similar to the seen classes.

%\subsection{Hierarchical sampling (HSamp)}

\subsection{Hierarchical ternary classification region proposal network (HTRPN)}

As mentioned in Section~\ref{sec: Find the potential proposals}, faster R-CNN has the ability to recognize a number of potential novel objects that belong to unseen classes, despite the fact that they are unlabelled during training on base class images. We hypothesize that this ability is because of the feature similarity between the features of some novel unseen classes and the base classes. As a result, the model would predict a novel unseen class object as a base class based on its resemblance. In other words, the novel objects contained by the negative anchors could probably have a relatively high classification score towards a base class that resembles them the most. We mark the negative anchors that contain potential novel unseen class objects as potential anchors ($A_p$), while others are marked as true negative anchors ($A_n^t$). Our goal is to distinguish between these two subsets.

Figure~\ref{fig:flowchart} visualizes our proposed architecture for improving FSOD. To better distinguish the set of $A_p$ from the set of $A_n^t$ during training, instead of performing binary classification to determine objectness in the original RPN, we propose a ternary objectness classification (i.e., $tobj\{{tobj_{pre}}^i, tobj_{gt}, iou_{gt}^a\}$, where ${tobj_{pre}}^i$ are the predicted ternary objectness scores between 0 and 1 for each class $i$; ground-truth value $tobj_{gt}=0$ indicates non-object, $tobj_{gt}=1$ represents a true object, $tobj_{gt}=2$ represents potential objects from unseen novel classes) so that potential objects that belong to unseen novel classes can be classified as a separate class, as visualized in Figure.~\ref{fig:flowchart-b}. For a training image, after the hierarchical sampling of the coarse negative anchors, we keep these negative anchors for any batch of anchors and perform instance-level sub-classification on them. Here, we set an instance-level classification threshold ($Thre_{cls}$). If we observe that the classification score is larger than the threshold  ($P(cls) > Thre_{cls}$) for a base class, then we set the anchor as an objectness-positive anchor, and mark its objectness loss with the ground truth of the label 2. For example the motorbike in Figure.~\ref{fig:flowchart-b}, the blue box is the ground truth box, active anchors are in green boxes, while negative anchors are in red boxes. Features of the negative anchors are sent to the RoI pooling layer to see if they could be predicted as a visually similar seen category (e.g. the negative anchor \ding{172} is predicted as base class ``bicycle'', then it is assigned with $tobj_{gt}=2$; but for anchor \ding{173} is kept as $tobj_{gt}=0$ since it does not pass the $Thre_{cls}$. We argue that our novel architecture will have a higher FSOD performance.

\begin{figure}[t]
    \centering
    \includegraphics[width=55mm]{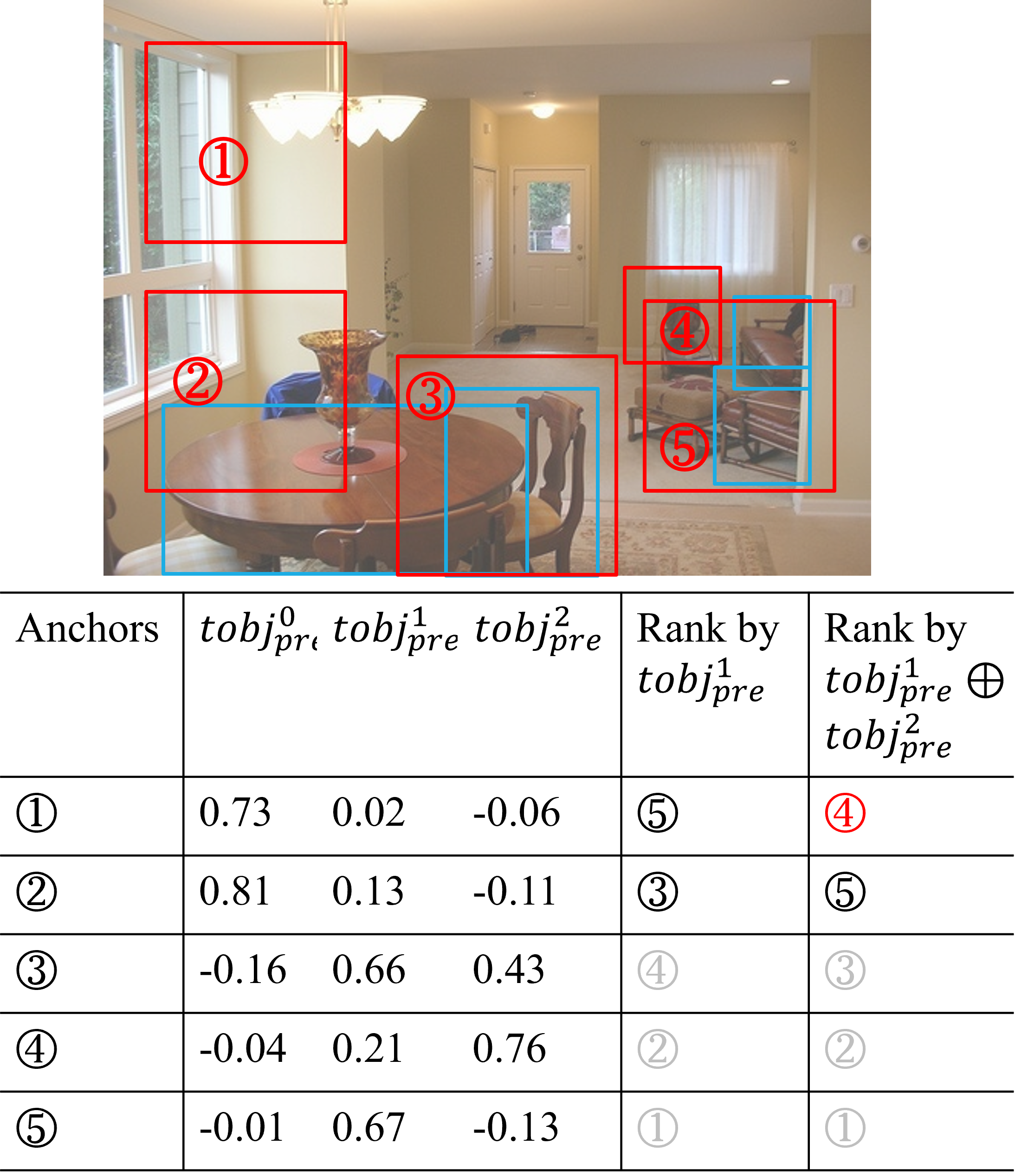}
    \caption{ \small The 4 blue boxes are the ground truth boxes and the 5 red boxes are the proposals. The proposal \ding{175} contains an unlabeled novel object chair. The top 2 proposals that are ranked by ${tobj_{pre}}^1 \oplus {tobj_{pre}}^2$ could successfully include proposal \ding{175}.}
    \label{fig:combine}
\end{figure}

In addition, we need customized solutions for the pre-training and fine-tuning when using our proposed HTRPN. Considering computational resource limitations, only the top 1000 proposals are used for RoI pooling in traditional RPN. Proposals are ranked by their objectness score of ${tobj_{pre}}^1$ during the pre-training stage since the model only learns to identify the base classes in this stage. However, in the fine-tuning stage, the ${tobj_{pre}}^1$ and ${tobj_{pre}}^2$ are both considered for ranking the proposals, because the objectness of some labeled objects might be predicted as ${tobj_{pre}}^2$ due to knowledge transfer from the pre-training stage. This step is crucial to realize the objectness consistency because the combination of ${tobj_{pre}}^1$ and ${tobj_{pre}}^2$ could represent the highly confident proposal and especially improve the possibility of determining positive anchors while inferencing. As shown in Figure~\ref{fig:combine}, if the top two proposals out of the five proposals are ranked only using ${tobj_{pre}}^1$, then the proposal \ding{175} would be ignored. However, when the top two proposals are ranked by ${tobj_{pre}}^1 \oplus {tobj_{pre}}^2$ (operator $\oplus$ could be addition or maximum, here we use maximum, details about $\oplus$ will be discussed in Appendix), the proposal \ding{175} could be correctly included. Such a scheme significantly increases the possibility to dig up the true object as much as possible. As a result, the anchors that contain potential novel unseen class objects are well distinguished from the coarse negative anchors. The ternary RPN will let the model maintain its sensitivity to identify new objects from classes that have never been seen before. In practice, not all potential objects that exist in training datasets are going to be singled out during training and only a subset of them could be found. However, these identified novel unseen class objects still can alleviate the confusion of the model during few-shot learning due to their dissimilarity to the seen classes.

\subsection{Contrastive learning on objectness}

To further increase the inter-class distances between $A_a$, $A_n^t$, and $A_p$ subsets in HTRPN, we also include an objectness contrastive learning head (ConsObj) in our architecture. Inspired by the existing literature \cite{Sun21,Khosla20}, the cropped features of proposals are sent into an encoder with their ground truth objectness logits to perform contrastive learning. The features of proposals are encoded as a default 128-dimension feature vector, and then the cosine similarity scores are measured between every two proposals. In this way, the HTRPN would give a higher objectness score.

\subsection{Training Loss}

The global total loss is composed of the classification loss ($\mathcal L_{Cls}$), the bounding box regression loss ($\mathcal L_{Bbox}$), our ternary objectness loss ($\mathcal L_{Tobj}$), and the RoI feature contrastive loss $\mathcal L_{Contra}$, as described in Equation.~\ref{eq:loss 1}. The $\mathcal L_{Contra}$ is computed using the contrastive head as described in FSCE ~\cite{Sun21}. We set $\alpha = 0.5$ to be the fixed weight for balancing the contrastive learning loss. 

\begin{equation}
\mathcal L = \mathcal L_{Cls} + \mathcal L_{Bbox} + \mathcal L_{Tobj} + \alpha \mathcal L_{Contra}
\label{eq:loss 1}
\end{equation}

Our proposed ternary objectness loss $\mathcal L_{Tobj}$ in Equation.~\ref{eq:loss 2} is a sum of the cross entropy objectness loss ($\mathcal L_{Obj}$) and ternary RPN feature contrastive learning loss ($\mathcal L_{Tcon}$). Similar to $\alpha$ in Equation.~\ref{eq:loss 1}, $\lambda$ is a balancing factor that is set to be equal 0.5 in our experiments.  

\begin{equation}
\mathcal L_{Tobj} = \mathcal L_{Obj} + \lambda \mathcal L_{Tcon}
\label{eq:loss 2}
\end{equation}

The ternary RPN contrastive learning loss $\mathcal L_{Tcon}$, is defined as an arithmetic mean of the weighted supervised contrastive learning loss $\mathcal L_{z_i}$ as the following:
\begin{equation}
\mathcal L_{Tcon}=\frac{1}{N_{Prop}}\sum_{i=1}^{N_{Prop}}w(iou_{gt}^p)\cdot\! \mathcal L_{z_i},
\label{eq:loss 3}
\end{equation}
 where $N_{Prop}$ represents the number of RPN proposals. Weights $w(iou_{gt}^p)$ are assigned by the function $g(*)$:
\begin{equation}
w(iou_{gt}^p) = \mathcal I\{iou_{gt}^p\geq \phi\}\cdot g(iou_{gt}^p),
\label{eq:loss 4}
\end{equation}
where $g(*)=1$ is a good hard-clip ~\cite{Sun21} and $\mathcal I\{*\}$ is a cut-off function that is   1 when $iou_{gt}^p\geq \phi\ $, otherwise is 0.

$\mathcal L_{z_i}$ in the RPN proposal contrastive learning loss is given as:
{\small
\begin{equation}
\mathcal L_{z_i}=\!\frac{-1}{N_{obj_{gt}^i}\!-\!1}\!\sum_{j=1,j\neq i}^{N_{Prop}}\!\mathcal I\{obj_{gt}^i= obj_{gt}^j\}\cdot\log\!\frac{e^{\tilde{z_i} \cdot\!\tilde{z_j}\!/\!\tau}}{\sum\limits_{k=1}\limits^{N_{Prop}}\!\mathcal I_{k\neq i}\cdot e^{\tilde{z_i}\cdot\tilde{z_k}/\!\tau}},
\label{eq:loss 5}
\end{equation}
}
where $z_i$ denotes the contrastive feature,  $obj_{gt}^i$ denotes the ground truth ternary objectness label for the $i$-th proposal,   $\tilde{z_i}$ denotes normalized features while measuring the cosine distances, and $N_{obj_{gt}^i}$ denotes the number of proposals with the same objectness label as $obj_{gt}^i$.

\begin{table*}[ht]
	\centering
	\resizebox{\textwidth}{!}{
		\begin{tabular}{ll|l|lllll|lllll|lllll}
			\toprule
			\multicolumn{2}{l|}{\multirow{2}{*}{\diagbox[height=25pt,innerrightsep=37pt]{Method}{Shot}}} & \multirow{2}{*}{Backbone} & \multicolumn{5}{c|}{Split1} & \multicolumn{5}{c|}{Split2} & \multicolumn{5}{c}{Split3} \\
			& & & 1 & 2 & 3 & 5 & 10 & 1 & 2 & 3 & 5 & 10 & 1 & 2 & 3 & 5 & 10\\
			\midrule
			LSTD & AAAI 18 ~\cite{Chen18} & \multirow{2}{*}{VGG-16} & 8.2 & 1.0 & 12.4 & 29.1 & 38.5 & 11.4 & 3.8 & 5.0 & 15.7 & 31.0 & 12.6 & 8.5 & 15.0 & 27.3 & 36.3 \\
			YOLOv2-ft & ICCV19 ~\cite{Wang19} & & 6.6 & 10.7 & 12.5 & 24.8 & 38.6 & 12.5 & 4.2 & 11.6 & 16.1 & 33.9 & 13.0 & 15.9 & 15.0 & 32.2 & 38.4 \\
			\midrule
			RepMet & CVPR 19 ~\cite{Karlinsky19} & InceptionV3 & 26.1 & 32.9 & 34.4 & 38.6 & 41.3 & 17.2 & 22.1 & 23.4 & 28.3 & 35.8 & 27.5 & 31.1 & 31.5 & 34.4 & 37.2 \\
			\midrule
			FRCN-ft & ICCV19 ~\cite{Wang19} & \multirow{12}{*}{FRCN-R101} & 13.8 & 19.6 & 32.8 & 41.5 & 45.6 & 7.9 & 15.3 & 26.2 & 31.6 & 39.1 & 9.8 & 11.3 & 19.1 & 35.0 & 45.1 \\
			FRCN+FPN-ft & ICML 20 ~\cite{Wang20} & & 8.2 & 20.3 & 29.0 & 40.1 & 45.5 & 13.4 & 20.6 & 28.6 & 32.4 & 38.8 & 19.6 & 20.8 & 28.7 & 42.2 & 42.1 \\
			TFA w/ fc & ICML 20 ~\cite{Wang20} & & 36.8 & 29.1 & 43.6 & 55.7 & 57.0 & 18.2 & 29.0 & 33.4 & 35.5 & 39.0 & 27.7 & 33.6 & 42.5 & 48.7 & 50.2 \\
			TFA w/ cos & ICML 20 ~\cite{Wang20} & & 39.8 & 36.1 & 44.7 & 55.7 & 56.0 & 23.5 & 26.9 & 34.1 & 35.1 & 39.1 & 30.8 & 34.8 & 42.8 & 49.5 & 49.8 \\
			MPSR & ECCV 20 ~\cite{Wu20} & & 41.7 & - & 51.4 & 55.2 & 61.8 & 24.4 & - & 39.2 & 39.9 & 47.8 & 35.6 & - & 42.3 & 48.0 & 49.7 \\
			Retentive R-CNN & CVPR 21 ~\cite{Fan21} & & 42.4 & 45.8 & 45.9 & 53.7 & 56.1 & 21.7 & 27.8 & 35.2 & 37.0 & 40.3 & 30.2 & 37.6 & 43.0 & 49.7 & 50.1 \\
			FSCE & CVPR 21 ~\cite{Sun21} & & 44.2 & 43.8 & 51.4 & \textcolor{blue}{61.9} & 63.4 & 27.3 & 29.5 & \textcolor{blue}{43.5} & 44.2 & 50.2 & \textcolor{blue}{37.2} & \textcolor{blue}{41.9} & \textcolor{blue}{47.5} & \textcolor{blue}{54.6} & \textcolor{blue}{58.5} \\
			TIP & CVPR 21 ~\cite{Li21a} & & 27.7 & 36.5 & 43.3 & 50.2 & 59.6 & 22.7 & 30.1 & 33.8 & 40.9 & 46.9 & 21.7 & 30.6 & 38.1 & 44.5 & 50.9 \\
			DC-Net & CVPR 21 ~\cite{Hu21a} & & 33.9 & 37.4 & 43.7 & 51.1 & 59.6 & 23.2 & 24.8 & 30.6 & 36.7 & 46.6 & 32.3 & 34.9 & 39.7 & 42.6 & 50.7 \\
			FSOD-UP & ICCV 21 ~\cite{Wu21b} & & 43.8 & \textcolor{red}{47.8} & 50.3 & 55.4 & 61.7 & \textcolor{blue}{31.2} & \textcolor{blue}{30.5} & 41.2 & 42.2 & 48.3 & 35.5 & 39.7 & 43.9 & 50.6 & 53.5 \\
			CME & CVPR 21 ~\cite{Li21b} & & 41.5 & \textcolor{blue}{47.5} & 50.4 & 58.2 & 60.9 & 27.2 & 30.2 & 41.4 & 42.5 & 46.8 & 34.3 & 39.6 & 45.1 & 48.3 & 51.5 \\
			KFSOD & CVPR 22 ~\cite{Zhang22} & & \textcolor{blue}{44.6} & - & \textcolor{red}{54.4} & 60.9 & \textcolor{red}{65.8} & \textcolor{red}{37.8} & - & 43.1 & \textcolor{red}{48.1} & \textcolor{blue}{50.4} & 34.8 & - & 44.1 & 52.7 & 53.9 \\
			\midrule
         Ours & & FRCN-R101 & \textcolor{red}{47.0} & 44.8 & \textcolor{blue}{53.4} & \textcolor{red}{62.9} & \textcolor{blue}{65.2} & 29.8 & \textcolor{red}{32.6} & \textcolor{red}{46.3} & \textcolor{blue}{47.7} & \textcolor{red}{53.0} & \textcolor{red}{40.1} & \textcolor{red}{45.9} & \textcolor{red}{49.6} & \textcolor{red}{57.0} & \textcolor{red}{59.7} \\
			\bottomrule
		\end{tabular}
    }
	\caption{The novel classes of nAP50fpr the PASCAL VOC dataset are evaluated on three different category splits with 1 to 10-shot scenarios.  The \textcolor{red}{highest score} of each few-shot setting is in red color, and the \textcolor{blue}{second highest score} is in blue color.}
	\label{tab:voc}
\end{table*}

\section{Experimental Results}

We empirically demonstrate that our proposed architecture and training procedure improve the FSOD performance. Our implementation is available as an Appendix.

\subsection{Expeiremntal Setup}
We use the Faster R-CNN as our object detection model and use ResNet-101 as the backbone along with the feature pyramid network (FPN). The evaluation scheme strictly follows the same paradigm as described in TFA~\cite{Wang20}. The mAP50 evaluation results are separately calculated on the base classes (bAP50) and the novel seen class (nAP50). We report our results on the PASCAL VOC and COCO datasets. The contrastive learning head in the fine-tuning stage is computed similarly to FSCE~\cite{Sun21}. We used four GPUs for training. The optimizer is fixed as SGD and the weight decay is 1e-4 with momentum as 0.9. We set our batch size equal to 16 for all experiments. The $Thre_{cls}$ is fixed as 0.75. These hyperparameters are not fine-tuned. In the pre-training stage, the top 1000 proposals used for RoI pooling, are ranked by the second objectness logit (is an object). While in the fine-tuning stage, the top 1000 proposals are ranked by the maximum of the second and the third objectness logits (potential object). There are many existing FSOD methods. We compare our performance against a subset of recently developed SOTA FSOD methods.% but also give the ablation study of our hyper-parameters.

\begin{table}[t]
	\centering
	\resizebox{0.5\textwidth}{!}{
		\begin{tabular}{ll|ll|ll}
			\toprule
			\multicolumn{2}{l|}{\multirow{2}{*}{\diagbox[innerrightsep=30pt]{Method}{Shot}}}&\multicolumn{2}{c|}{Novel AP}&\multicolumn{2}{c}{Novel AP75}\\
			&&10&30&10&30\\
			\midrule
			TFA w/ cos&ICML20~\cite{Wang20}&10.0&13.7&9.3&13.4\\
			FSCE&CVPR21~\cite{Sun21}&11.9&\textcolor{blue}{16.4}&\textcolor{blue}{10.5}&\textcolor{blue}{16.2}\\
			SRR-FSD&CVPR21~\cite{Zhu21} &11.3&14.7&9.8&13.5\\
			SVD&NeurIPS21~\cite{Wu21a}&\textcolor{blue}{12.0}&16.0&10.4&15.3\\
			FORD+BL&IMAVIS22~\cite{Nguyen22}&11.2&14.8&10.2&13.9\\
			N-PME&ICASSP22~\cite{Liu22}&10.6&14.1&9.4&13.6\\
			\midrule
			Our& &\textcolor{red}{12.1}&\textcolor{red}{17.2}&\textcolor{red}{11.2}&\textcolor{red}{17.1}\\
			\bottomrule
		\end{tabular}
	}
	\caption{Evaluation on COCO dataset for novel classes for AP and AP75 settings. The \textcolor{red}{highest score} of each few-shot setting is in red, and the \textcolor{blue}{second highest score} is in blue.}
	\label{tab:coco}
\end{table}

\subsection{Results on PASCAL VOC}
For the PASCAL VOC 2007 and 2012 dataset, 15 categories are chosen as the base classes for pre-training, and the remaining 5 categories served as the novel classes. We follow the 3 different categories splits defined in TFA ~\cite{Wang20}. To achieve a fairer comparison, TFA ~\cite{Wang20} defined three kinds of combinations of base classes and novel classes, namely split1, split2, and split3. In each split, we evaluate the average precision for novel classes (nAP) on 1,2,3,5,10 shots separately. The training iterations are 8000 for each training epoch. We set the initial learning rate to 0.02. Our results on the three categories splits are reported in Table.~\ref{tab:voc}. We observe that our proposed method reaches new SOTA performance in most cases. Especially, our method is more effective when the $n$-shot is smaller.

\subsection{Results on COCO}
For the COCO dataset, 60 categories are selected as base classes, and the remaining 20 categories are served as novel classes. The training iterations are set to 20000 during the training stage with an initial learning rate of 0.01. AP for novel classes is evaluated upon $n=10$ and $n=30$ shots separately. Our experiment results for COCO are shown in Table.~\ref{tab:coco}. We again observe that our method outperforms the previous SOTA works in all cases and in some cases the margin of improvement is significant. These experiments demonstrate that our method is effective.

\subsection{Ablation Study}

Firstly, we discuss the effectiveness of our proposed modules separately, including the Hierarchical sampling of the RPN, the ternary objectness classification, and the contrastive head of the objectness. We implemented the ablation study experiment on PASCAL VOC 5-shot scenario. Each proposed module is added to the original network in an accumulated manner. The results are presented in Table.~\ref{tab:ablation1}. We observe that all our proposed modules are necessary for optimal performance. By adding the HSamp, we can see that a balanced sampling in RPN is necessary, as it provides comprehensive improvement of $bAP$ and $nAP$ during the pre-training and the fine-tuning stages. We can also observe the results of adding the ternary objectness module indicate that our method will further improve the $nAP$ and do no significant harm to the $bAP$. While the contrastive objectness part demonstrated that it is a simple yet effective way to help build a stronger RPN that could further improve the $bAP$ and $nAP$.

\begin{table}[t]
	\centering
	\resizebox{0.4\textwidth}{!}{
	\begin{tabular}{l|l|l|l}
		\toprule
		  Modules & \makecell[l]{bAP\\(pre-trained)} & \makecell[l]{bAP\\(fine-tuned)} & \makecell[l]{nAP\\(fine-tuned)} \\
		\midrule
		FSCE Baseline* & 80.5 & \textbf{68.9} & 57.2 \\
		+ HSamp & \textbf{80.7} & \textbf{68.9} & 57.6 \\
		+ Ternary Objectness & 78.5 & 67.8 & 61.9 \\
		+ Contrastive Objectness & 78.9 & 68.6 & \textbf{62.9} \\
		\bottomrule
	\end{tabular}
    }
	\caption{Ablation studies on different modules. The effect of incrementally adding each module to the Baseline network is demonstrated respectively. Sign * represents our reproductive results. We listed the base class mAP50 (bAP) during the pre-training and fine-tuning stage, as well as the novel class mAP50 (nAP) during the fine-tuning stage.}
	\label{tab:ablation1}
\end{table}

Additionally, we study the influence of different hyperparameter $Thre_{cls}$ settings. We use five $Thre_{cls}$ values from 0.05 to 0.95 for training and record the $nAP$ accordingly, as shown in Table.~\ref{tab:ablation2}. We observe that for lower $Thre_{cls}$, more candidate potential novel proposals can be distinguished. However, we have lower confidence and consequently lower quality. However, when a higher threshold $Thre_{cls}$ is used, the number of candidates for potential novel proposals is smaller, which is insufficient to optimize objectness in our framework. As the result indicates, $Thre_{cls} = 0.75$ is a reasonable value for filtering the candidate proposals relatively well. For additional quantitative analysis please refer to the supplementary materials.

\begin{table}[t]
	\centering
	\resizebox{0.3\textwidth}{!}{
	\begin{tabular}{l|l|l|l|l|l}
		% \toprule
		% $Thre_{cls}$ & nAP \\
		% \midrule
		% 0.05 & 60.5 \\
		% 0.25 & 61.2 \\
		% 0.5 & 62.1 \\
		% 0.75 & \textbf{62.9} \\
		% 0.95 & 61.4 \\
		% \bottomrule
        \toprule
		$Thre_{cls}$ & 0.05 & 0.25 & 0.5 & 0.75 & 0.95\\
		\midrule
		nAP & 60.5 & 61.2 & 62.1 & \textbf{62.9} & 61.4\\
		\bottomrule
	\end{tabular}
 }
	\caption{Ablation studies on different hyperparameter settings. The effect of adjusting the $Thre_{cls}$ is demonstrated. The highest nAP has been bolded.}
	\label{tab:ablation2}
\end{table}

\section{Conclusions}

 We improved the quality FSOD using R-CNN-based architecture via studying the phenomenon of objectness inconsistency due to the potential unlabeled novel objects that belong to unseen classes. By a balance anchor sampling strategy, we enhance the possibility of identifying anchors that may contain objects from unseen classes. In addition, we proposed   HTRPN which leads to the recognition ability of potential novel objects and further enhances the objectness consistency. Our method mitigates model confusion about the novel classes and achieves SOTA performance on standard datasets. Future works include extensions to identify novel unseen classes in a zero-shot learning setting.

{\small
\bibliographystyle{ieee_fullname}
\bibliography{egbib}
}
% \end{document}

\newpage
\appendix

We include complementary qualitative and quantitative analyses of our method in this Appendix.

%%%%%%%%% BODY TEXT
\section{Ranking proposals}

As mentioned in our main paper Section 4.2, the top 1,000 proposals that are most likely to contain objects are selected for the training according to their objectness logit scores. Since our proposed hierarchical ternary region proposal network (HTRPN) has three objectness scores (${tobj_{pre}}^0$ indicates the predicted score of non-object, ${tobj_{pre}}^1$ represents the score of the true object, ${tobj_{pre}}^2$ represents the score of potential objects from unseen novel classes), we rank the proposals by combing their ${tobj_{pre}}^1$ and ${tobj_{pre}}^2$, mark as ${tobj_{pre}}^1 \oplus {tobj_{pre}}^2$. Operator $\oplus$ could be either arithmetical addition (${tobj_{pre}}^1 + {tobj_{pre}}^2$) or maximum ($max({tobj_{pre}}^1, {tobj_{pre}}^2)$).

\begin{figure}[t]
    \centering
    \includegraphics[width=80mm]{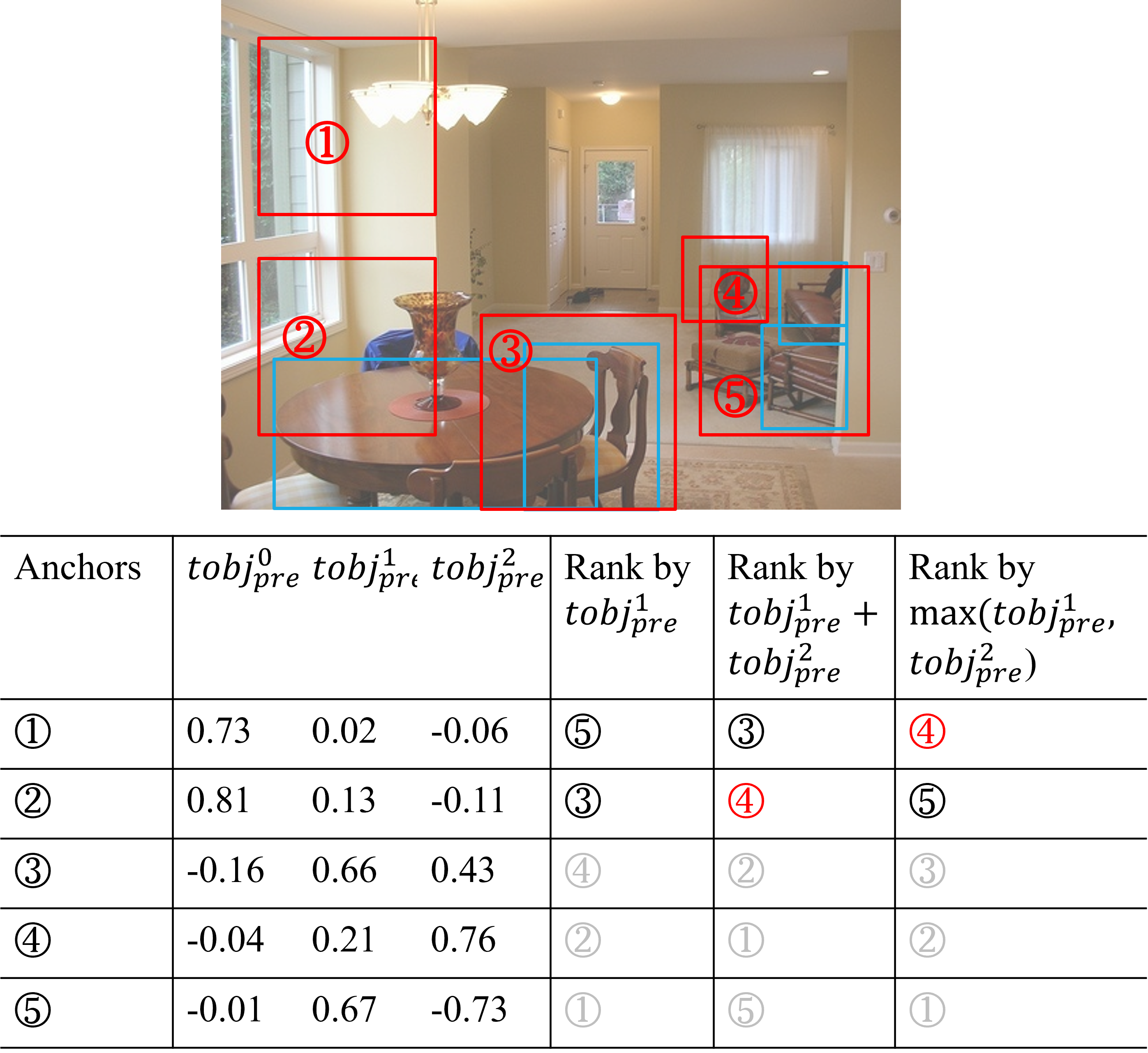}
    \caption{The 4 blue boxes are the ground truth boxes and the 5 red boxes are the proposals. The proposal \ding{175} contains an unlabeled novel object chair. The top 2 proposals that are ranked by ${tobj_{pre}}^1 \oplus {tobj_{pre}}^2$ could successfully include proposal \ding{175}.}
    \label{fig:maximum}
\end{figure}

Our examples in Figure.~\ref{fig:maximum} demonstrate the difference between these two types of the operator $\oplus$. In the image, labeled base class objects (tables and chairs) are in blue boxes. For five exemplary proposals \ding{172} to \ding{176} in red boxes with their ternary objectness logit scores, we intend to pick the top 2 of them. If we rank the proposals only by ${tobj_{pre}}^1$, then the proposals that contain unlabeled potential objects (e.g. proposal \ding{175}) will more likely be ignored since they are trained to have higher ${tobj_{pre}}^2$ score instead of ${tobj_{pre}}^1$. Therefore, it is necessary to take the ${tobj_{pre}}^2$ score into account, and according to our HTRPN, the objectness of an object should be presented by ${tobj_{pre}}^1$ and ${tobj_{pre}}^2$ together so that the rank of the potential proposal such as \ding{175} could be significantly promoted. However, by the method of ${tobj_{pre}}^1 + {tobj_{pre}}^2$, the rank of some proposals will be negatively influenced by the negative logit values, such as the proposal \ding{176}. On the contrary, by applying the maximum between ${tobj_{pre}}^1$ and ${tobj_{pre}}^2$, the rank of proposal \ding{176} will not be degraded by the singular negative values.

\section{Magnitude of potential novel objects}

\begin{figure}[t]
    \centering
    \begin{subfigure}{\linewidth}
        \centering
    	\includegraphics[width=85mm]{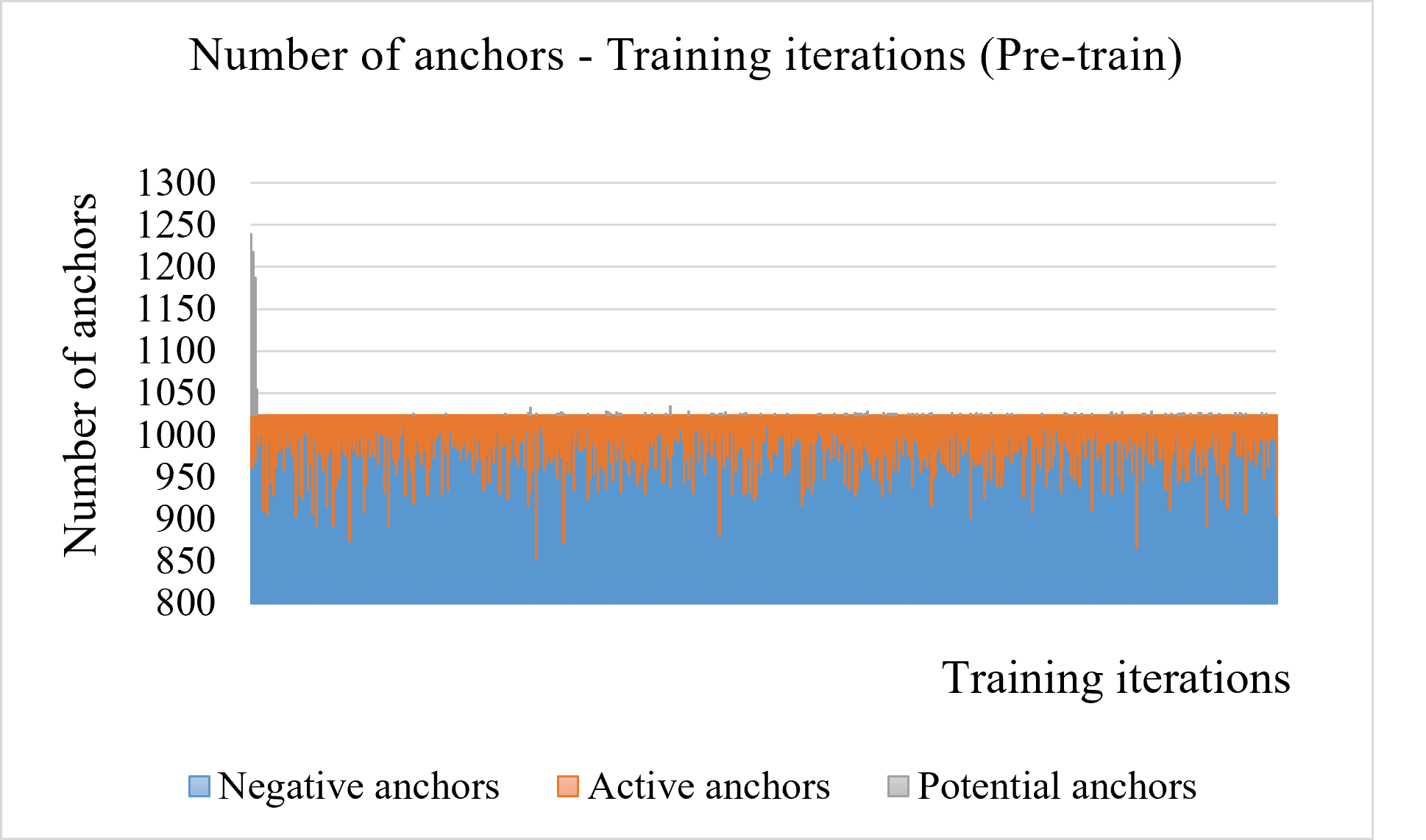}
    	\caption{Number of anchors - Training iterations chart of the pre-training stage.}
    	\label{fig:number of anchors pre train}
    \end{subfigure}
	\centering
    \begin{subfigure}{\linewidth}
        \centering
    	\includegraphics[width=85mm]{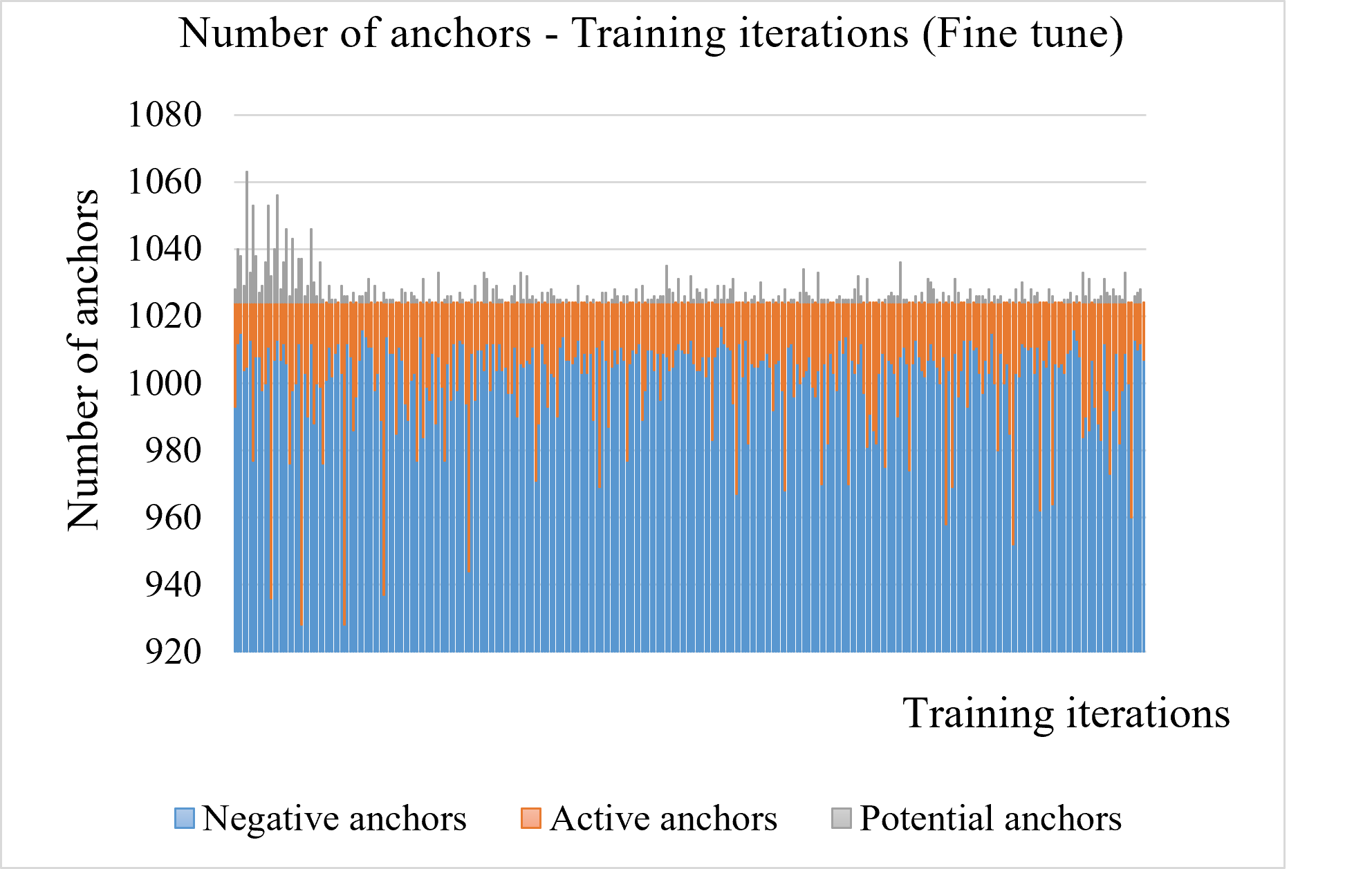}
    	\caption{Number of anchors - Training iterations chart of the fine-tuning stage.}
    	\label{fig:number of anchors fine tune}
    \end{subfigure}
	\caption{The horizontal axis is the training iterations during fine-tuning, and the vertical axis indicates the number of anchors of each training iteration.}
	\label{fig:number of anchors}
\end{figure}

We record the number of potential objects during the pre-training and fine-tuning stages to demonstrate the statistical concept of the potential objects. Our experiments are implemented on the PASCAL VOC dataset with the 5-shot setting. The instance-level ternary classification threshold is fixed as $Thre_{cls}=0.75$. The relation between the number of anchors and the training iteration in the pre-training stage is shown in Figure.~\ref{fig:number of anchors pre train}, the overall training anchors for each image is 256, as we claimed in~\ref{sec: Find the potential proposals}, thus, for a batch size of 16 on 4 GPUs, the overall training anchors on a single GPU would be around $(16/4)*256=1024$. The negative anchors (blue bars) are the majority, and the number of active anchors (orange bars) is around 0 to 50 for each iteration, which is about 4\% of all anchors. However, the number of potential anchors (gray bars) converges with training iterations and stays at around 0 to 5 for each iteration, which indicates our model is getting more stable for recognizing the potential novel objects. Furthermore, as shown in Figure.~\ref{fig:number of anchors fine tune}, during 5-shot fine-tuning, the trend of the potential anchors is as same as in the pre-training stage. However, when transferring to novel classes, the pre-trained model tends to predict some labeled novel objects as potential novel objects and the beginning of the fine-tuning stage, which could be considered as the inertia of the pre-trained model, therefore, the number of potential anchors is relatively high in the first several iterations of training. This also supports our theory that all true/potential positive ternary objectness of anchors should be presented by the combination of ${tobj_{pre}}^1$ and ${tobj_{pre}}^2$. And then the number of potential anchors is gradually stabilized to around 0 to 15 for each iteration, which is about 1.5\% of all anchors in an iteration.

\begin{table}[t]
	\centering
	\resizebox{0.45\textwidth}{!}{
	\begin{tabular}{l|l|l}
		\toprule
		  Hyperparameters & PASCAL VOC & COCO \\
		\midrule
		Learning rate & 0.01 & 0.001 \\
		Weight decay & 1e-4 & 1e-4 \\
		Optimizer & SGD & SGD \\
		Temperature & 0.2 & 0.2 \\
		Contrastie loss weight & 0.5 & 0.5 \\
		$Thre_{cls}$ & 0.75 & 0.75 \\
		\bottomrule
	\end{tabular}
    }
	\caption{Hyper parameters that we use for experiments.}
	\label{tab:hyperparameters}
\end{table}

\begin{figure}[t]
    \centering
    \includegraphics[width=80mm]{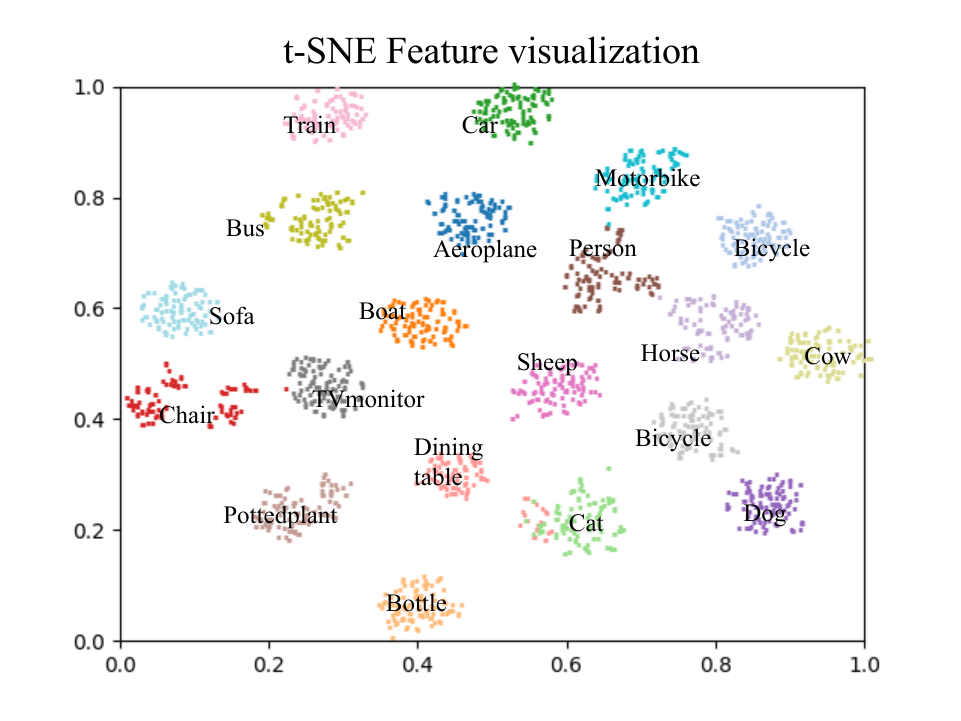}
    \caption{The t-SNE visualization of our HTRPN.}
    \label{fig:tsne}
\end{figure}

The t-SNE visualized clustering for the 5-shot scenario of the PASCAL VOC dataset is in Figure.~\ref{fig:tsne}, each point is an object feature, and the features from the same categories are more likely to form a cluster. A denser cluster indicates a smaller intra-class distance and better recognition ability of its category.

\section{Implemented details}

We use 4 Nvidia T4 Tensor Core GPUs for both training and evaluation of all datasets. The general hyperparameters we used are listed in Table.~\ref{tab:hyperparameters}. As we are applying contrastive learning for RPN and RoI pooling, and the feature space of the pre-training and fine-tuning samples are totally different, the weights of these contrastive learning modules are not transferable from the pre-trained model to the fine-tuning tasks. Therefore, we remove the weights from the pre-trained model for fine-tuning training. The weights from the Resnet backbone and RoI pooling are unfrozen during fine-tuning.

% {\small
% \bibliographystyle{ieee_fullname}
% \bibliography{egbib}
% }

\end{document}